\documentclass[preprint,authoryear,11pt,times]{elsarticle}

\usepackage{lineno}
\usepackage{graphicx}
\usepackage{epstopdf}
\usepackage{amssymb}
\usepackage{amsfonts,amsmath,bm,bbm}
\usepackage{mathtools}
\usepackage{enumerate}
\usepackage{indentfirst}
\usepackage[font=footnotesize]{caption}
\usepackage{floatrow}
\usepackage{enumitem}
\usepackage[table,xcdraw,dvipsnames]{xcolor}
\usepackage{url}
\usepackage{multirow}
\usepackage{cite}
\usepackage{floatrow}
\usepackage{booktabs} 
\usepackage{placeins}
\usepackage{multicol}
\usepackage{enumitem}
\usepackage{hyperref}
\usepackage[margin=2.5cm]{geometry}
\DeclareMathOperator{\argmin}{argmin}

\usepackage{orcidlink}
\usepackage{setspace}

\usepackage{tikz}
\usepackage{varwidth}
\usetikzlibrary{patterns,decorations.pathreplacing}

\journal{International Journal of Forecasting}

\begin{document}
\onehalfspacing
\begin{frontmatter}

\title{Isotonic Quantile Regression Averaging for uncertainty quantification of electricity price forecasts}

\author[szd]{Arkadiusz Lipiecki \orcidlink{0000-0003-1118-0388}} %% Author name
\ead{arkadiusz.lipiecki@pwr.edu.pl}
 \affiliation[szd]{organization={Department of Computational Social Science, Faculty of Management, Wroc{\l}aw University of Science and Technology},%Department and Organization
addressline={}, 
city={Wroc{\l}aw},
postcode={50-370}, 
country={Poland}}

\author[k43]{Bartosz Uniejewski \orcidlink{0000-0001-8563-2514} \corref{cor1}} %% Author name
\ead{bartosz.uniejewski@pwr.edu.pl}
\cortext[cor1]{Corresponding author: Bartosz Uniejewski}
\affiliation[k43]{organization={ Department of Operations Research and Business Intelligence, Wroc{\l}aw University of Science and Technology},%Department and Organization
addressline={}, 
city={Wroc{\l}aw},
postcode={50-370}, 
country={Poland}}
%% Abstract
\begin{abstract}
Quantifying the uncertainty of forecasting models is essential to assess and mitigate the risks associated with data-driven decisions, especially in volatile domains such as electricity markets. In this paper, we propose {a computationally efficient isotonic variant of} Quantile Regression Averaging (QRA), called Isotonic Quantile Regression Averaging (iQRA), for generating probabilistic forecasts from ensembles of point forecasts. {Building on QRA, we introduce stochastic order constraints as a hyperparameter-free regularization mechanism that reduces computational complexity while maintaining competitive forecast accuracy.} We further demonstrate that the same isotonic regularization principle can be applied to the smoothed estimator of QRA. In an extensive forecasting study of {two European day-ahead electricity markets}, we show that iQRA {and its smoothed variant} consistently rank among the top state-of-the-art postprocessing methods in terms of reliability and sharpness. It produces well-calibrated prediction intervals, {with smoothed iQRA} providing superior reliability even when compared against coverage-oriented conformal prediction. {We further introduce a battery arbitrage evaluation, showing that the leading probabilistic methods achieve the highest profit per trade and that probabilistic risk control improves the economic value of trading decisions.} {Isotonic regularization offers a scalable hyperparameter-free variable selection for linear quantile regression, making iQRA attractive for practical decision-support systems in energy markets.}

\end{abstract}


\end{frontmatter}


\section{Introduction}
The primary goal of a point forecasting model is to provide an accurate prediction of the future value of a variable of interest to aid in the decision making process \citep{pet:etal:22}. However, any model inherently produces predictions with error. Therefore, decisions based on artificial intelligence are subject to risk. To assess and mitigate this risk, we use uncertainty quantification techniques that allow us to learn and predict the distribution of model errors \citep{abd:etal:21}. Properly accounting for the uncertainty of forecasts is critical for operational decisions, especially in electricity markets \citep{wen:pin:25}, where supply must constantly match demand, leading to highly volatile prices~\citep{hon:pin:etal:20}.

In real-world scenarios, decision makers often use forecasts from multiple sources, sometimes provided by third parties, and therefore may not have access to or influence over the forecast generation process. For this reason, model agnostic postprocessing methods that use only the out-of-sample forecasts are attractive tools for supporting managerial decisions \citep{lip:uni:wer:24}. Furthermore, the popularity of machine learning model ensembles motivates the development of ensemble-based uncertainty quantification methods in electricity price forecasting~\citep{bru:etal:2025}.

From an operational research perspective, probabilistic forecasting is valuable because it allows for the optimization of decisions under uncertainty instead of basing them on a single expected scenario. In energy applications, probability forecasts support critical operational actions, such as reserve planning, dispatch decisions, and market bidding. The cost of adverse forecast errors in these areas can be substantial. Probabilistic forecasts allow decision-makers to explicitly balance risk and return in bidding strategies and optimize reserve levels in power systems \citep{tay:17}. Furthermore, probabilistic forecasts can be directly translated into decision rules by comparing forecast probabilities against thresholds derived from economic costs. This leads to rational, expected-cost decisions in practice \citep{tay:jeo:18}.

{The literature on probabilistic electricity price forecasting has considered a wide range of statistical and machine learning approaches, including ARX-GARCH models \citep{hic:etal:12, jan:puc:23, bil:gia:gro:rav:23}, conformal prediction \citep{kat:zie:21, zaf:etal:22}, generalized additive models for location, scale and shape \citep[GAMLSS;][]{ser:11,hir:GAMLSS:24}, and distributional deep neural networks \citep[DDNN;][]{mar:nar:wer:zie:23}. Despite this methodological diversity, maintaining reliable probabilistic forecasts under recent extreme market conditions remains challenging. In a recent empirical study, \citet{lip:uni:wer:24} show that the performance of even well-established models can vary substantially across different market periods.}

Recently, a new nonparametric method, called Isotonic Distributional Regression (IDR), has been proposed for estimating probabilistic distributions under a stochastic order constraint between the target random variable and the covariates~\citep{hen:zie:gne:21}. {When the covariates are point estimates of the same variable, the isotonicity of the distribution of the target is desirable and intuitive. Holding all else constant, predictive quantiles should not decrease when the estimated expected value increases. This observation motivates the use of IDR for postprocessing forecast ensembles into predictive distributions~\citep{hen:zie:gne:21, wal:etal2024}.} However, the performance of IDR as a stand-alone method for uncertainty quantification in electricity price forecasting has been rather disappointing~\citep{lip:uni:wer:24}. It was outperformed by standard approaches such as Conformal Prediction (CP) and Quantile Regression Averaging (QRA). Nevertheless, the isotonicity of the target with respect to its predictions is an attractive property that regularizes the solution of a distribution learning problem in an explicable and intuitive way. Therefore, we {adapt the idea of isotonic regularization to ensemble-based quantile regression methods and propose Isotonic Quantile Regression Averaging (iQRA), an isotonic variant of QRA. Furthermore, we show that the same principle can be applied to Smoothed Quantile Regression Averaging (SQRA), yielding its isotonic counterpart, Isotonic Smoothed Quantile Regression Averaging (iSQRA).} Our approach does not require any hyperparameters and can be easily implemented by adapting the linear programming formulation of standard quantile regression, {reducing its computational cost and yielding interpretable regularization that preserves the isotonic relationship between ensemble forecasts and quantile predictions.}

We emphasize that the isotonicity of quantile estimates is not the {novel} contribution of this paper and has been studied in various forms~\citep{cas:cry:1976, poi:tho:2000, bol:eil:aer:2006, tak:etal:2006, mos:dum:2020, can:2018}. However, despite the popularity of linear quantile regression in postprocessing {point forecasts, the isotonic regularization seems to have been overlooked in this application}. 

To provide a comprehensive analysis of {the proposed isotonic quantile regression methods}, we conduct an extensive study of {the German and French} day-ahead electricity markets using {ensembles of point forecasts generated by autoregressive neural networks (NARX) and eXtreme Gradient Boosting (XGB) models}. We compare {iQRA and iSQRA} with several state-of-the-art post-processing methods for uncertainty quantification \citep{lip:uni:wer:24}{, including their non-isotonic counterparts, QRA and SQRA}. Following the best practices of Lago et al. \citep{lago:etal:2021}, we {evaluate the competing methods over a six year out-of-sample period spanning 2020--2025, which includes the COVID-19 pandemic, the European energy crisis, and the subsequent market stabilization period}. {To assess the robustness of the proposed approaches, we consider ensembles consisting of 10, 25, and 50 members and analyze the sensitivity of the results to the underlying point forecasting model. The results demonstrate that isotonic regularization yields reliable and sharp probabilistic forecasts while reducing computational complexity and eliminating the need for hyperparameter tuning.} {Finally, we complement the statistical evaluation with a simple trading strategy to assess the economic value of the resulting probabilistic forecasts.} To provide {readers} with a ready-to-use implementation, we added iQRA {and iSQRA} to the open-source Julia package \textit{PostForecasts.jl}~\citep{lip:wer:2025}.

The rest of the paper is organized as follows. In Section \ref{sec:Dataset} we present the datasets, then in Section \ref{sec:Base} we explain how the point forecasts of day-ahead electricity prices are computed. Next, in Section \ref{sec:Postprocess} we describe the methods used to obtain probabilistic forecasts. In particular, we introduce the iQRA {and iSQRA} approaches. In Section \ref{sec:Results} we compare the performance of all considered methods in terms of both reliability and sharpness of the probabilistic forecasts {and evaluate their performance in a stylized trading application}. Finally, in Section \ref{sec:Conclusions} we summarize the main results.

\begin{figure*}[b!]
 \centering
 \includegraphics[width = 0.88\linewidth]{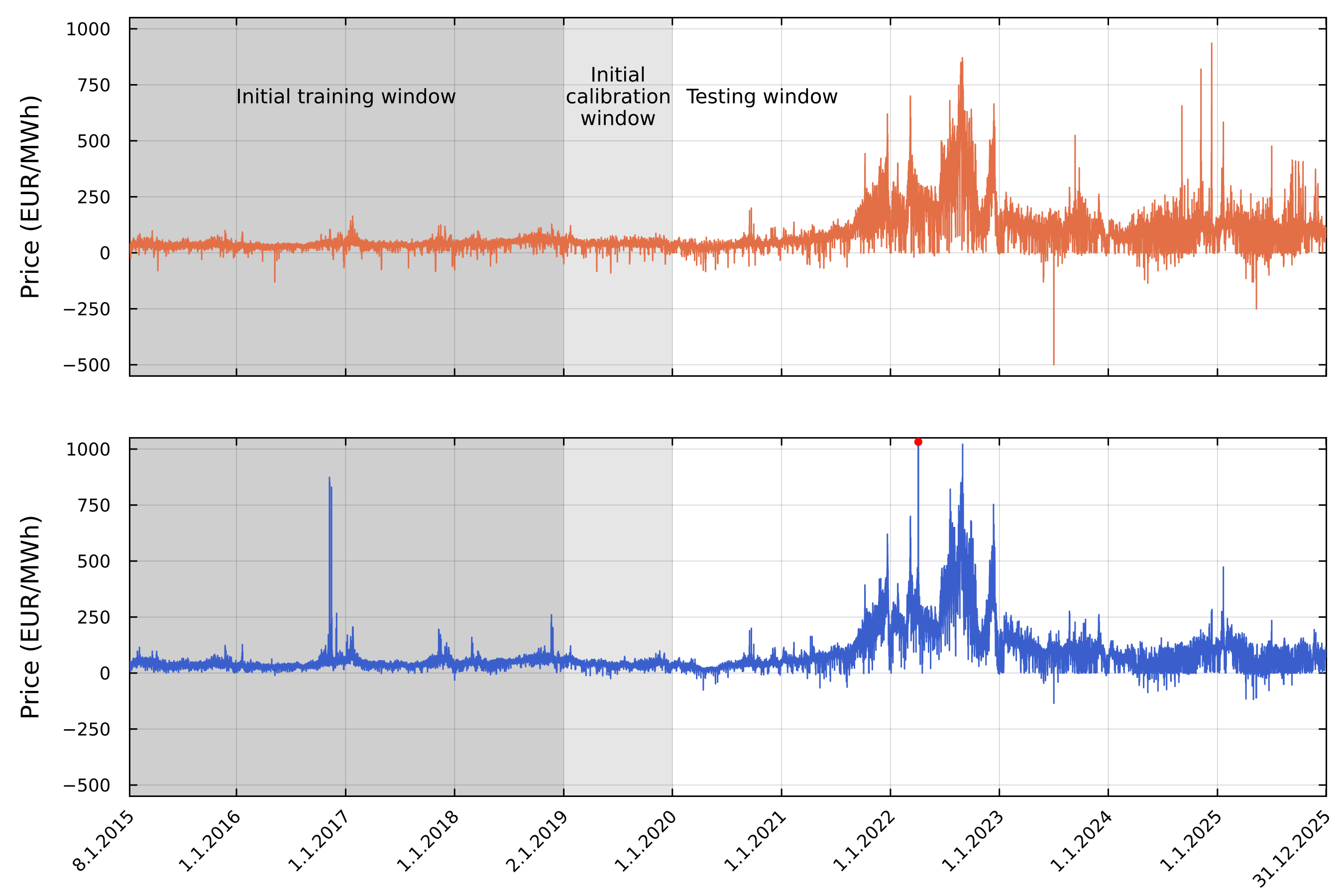}
 \caption{{Day-ahead electricity prices in the Germany-Luxembourg (top) and French (bottom) markets for the period 8.1.2015--31.12.2025. The first vertical dashed line marks the end of the 1456-day training window for the point forecasting models. The second dashed line marks the end of the 364-day calibration window for the post-processing methods and the beginning of the 2192-day out-of-sample test period. The red dot in the French market series indicates the peak price of 3000 EUR per MWh.}}

 \label{fig:data}
\end{figure*}

\section{Data and empirical framework}
\subsection{Test ground}
\label{sec:Dataset}

As a case study to demonstrate the effectiveness of the iQRA {and iSQRA} approaches, we focus on the {day-ahead European market coupling for Germany-Luxembourg and French} electricity market{s} -- {two of the largest and most dynamic energy systems in Europe}. To support this analysis, we have compiled {market-specific datasets} that reflect both the market structure and the key drivers of electricity prices. The core of the dataset consists of day-ahead electricity prices from the {ENTSO-E} transparency platform.\footnote{Note that prices refer to the Germany-Luxembourg bidding zone, but prior to October 2018 this zone also included Austria}. To account for supply and demand fundamentals, we included day-ahead forecasts of system load, solar generation, and aggregated wind generation (onshore and offshore) in {the corresponding country}, also from ENTSO-E.\footnote{Note that solar and wind generation have been combined into a single time series to reflect renewable energy generation.} Recognizing the influence of global energy markets on electricity prices, we have further enriched the dataset with commodity market indicators -- namely the closing prices of coal (API2), natural gas (TTF), crude oil (Brent), and carbon emission allowances (EUA) -- sourced from Investing.com.

The data collected was pre-processed to ensure consistency. Several variables - such as load and renewable generation - were initially available at 15 minute resolution. These were aggregated into hourly time series to ensure consistency of the dataset. Time shifts due to the transition between Central European Time (CET) and Central European Summer Time (CEST) were also taken into account. During the spring changeover to CEST, when an hour is skipped, missing values were imputed {with data from the previous hour}. Conversely, during the fall changeover to CET, when an hour is repeated, duplicate values were replaced by their arithmetic mean. {Additionally, since both markets moved to 15-minute market time units on 1.10.2025, the quarter-hourly prices for the last three months of the sample were aggregated to hourly values using an unweighted arithmetic mean. Hence, for this final quarter, the forecasted variable is the hourly average of four 15-minute day-ahead prices.}

All collected time series span from 8.1.2015 to 31.12.{2025}, with a {6-year} out-of-sample test period starting on 1.1.2020, as shown in Fig.~\ref{fig:data}. To obtain the forecasts, we employ a rolling window scheme. First, a 1456-day (208 weeks, approximately four years) rolling training window is used to generate point forecasts of electricity prices. Once these point predictions are available, a second 364-days rolling calibration window is used to estimate the postprocessing model\footnote{{As a sensitivity check, we conducted limited empirical tests and considered shorter and longer calibration windows of 112 and 728 days. The 364-day window yielded the lowest CRPS for all methods. The exact ranking changed for the shortest window, but the main qualitative conclusions remained unchanged.}} This two-step procedure enables dynamic recalibration and supports robust uncertainty quantification over time.

\subsection{Point forecasting}
\label{sec:Base}

\begin{figure}[b!]
	\tikzset{%
		input neuron/.style={
			circle,
			draw,
			% fill=blue!50,
			minimum size=.5cm
		},
		every neuron/.style={
		circle,
		draw,
		% fill=blue!50,
		minimum size=.6cm
		},
		neuron missing/.style={
			draw=none, 
			scale=1.2,
			text height=.25cm,
			execute at begin node=\color{black}$\vdots$
		},
		sigmoid/.style={path picture= {
				\begin{scope}[x=.7pt,y=7pt]
					\draw plot[domain=-6:6] (\x,{1/(1 + exp(-\x))-0.5});
				\end{scope}
			}
		},
		linear/.style={path picture= {
			\begin{scope}[x=5pt,y=5pt]
				\draw plot[domain=-1:1] (\x,\x);
			\end{scope}
		}
	},
	}

	\centering
	\begin{tikzpicture}[x=1.7cm, y=.65cm, >=stealth]
	
	\foreach \m/\l [count=\y] in {1,2,3,4,5,6,7,8,9,10,11,12,13}
	\node [input neuron/.try, neuron \m/.try] (input-\m) at (0,2.5-\y) {};

  \foreach \m/\l [count=\y] in {14}
	\node [input neuron/.try, neuron \m/.try] (input-\m) at (0,-12.5) {};
	
	\foreach \m [count=\y] in {1,2,3,4,5}
	\node [every neuron/.try, neuron \m/.try, fill= black!20, sigmoid ] (hidden-\m) at (2.,2.5-\y*2.5) {};
	
	\foreach \m [count=\y] in {1}
	\node [every neuron/.try, neuron \m/.try, fill= blue!20, linear ] (output-\m) at (4,-5) {};
	
	\foreach \l [count=\i] in {$p_{d-1,h}$,$p_{d-2,h}$,$p_{d-7,h}$,$p_{d-1,24}$,$p_{d-1}^{\min}$,$p_{d-1}^{\max}$,${Load}_{d,h}$, ${RES}_{d,h}$, $\text{Coal}_{d-2}^{\text{close}}$, $\text{Gas}_{d-2}^{\text{close}}$, $\text{Oil}_{d-2}^{\text{close}}$, $\text{EUA}_{d-2}^{\text{close}}$, $D_1$,$D_7$}
	\draw [<-] (input-\i) -- ++(-1.2,0)
	node [above, midway] {\l};

  \node[] at (-0.6,-11.5) {\ldots};
  
	\foreach \l [count=\i] in {1,2,3,4,5}
	\node [above] at (hidden-\i.north) {};
	
	\foreach \l [count=\i] in {1}
	\draw [->] (output-\i) -- ++(1.2,0)
	node [above, midway] {{$p_{d,h}$}};
	
	\foreach \i in {1,...,14}
	\foreach \j in {1,...,5}
	\draw [->] (input-\i) -- (hidden-\j);
	
	\foreach \i in {1,...,5}
	\foreach \j in {1}
	\draw [->] (hidden-\i) -- (output-\j);
	
	\foreach \l [count=\x from 0] in {Input, Hidden, Output}
	\node [align=center, above] at (\x*2,2) {\l \\[-3pt] {layer}};
	
	\end{tikzpicture}	
	\caption{Visualization of the NARX network with five hidden neurons with hyperbolic tangent activation functions and one linear output neuron.}
	\label{fig:NARX}
\end{figure}

Our baseline model for producing point forecasts of day-ahead electricity prices is the feedforward neural network, known as Nonlinear Autoregression with eXogenous variables (NARX). The aim of our paper is not to provide the best possible point forecasting model, but to propose and test a new method for uncertainty quantification. Therefore, the structure of our neural networks is directly adapted from existing studies on electricity price forecasting \citep{mar:uni:wer:2020}. The NARX model is thus a shallow neural network with 5 neurons and hyperbolic tangent activation functions in a single hidden layer, and a linear function in the output layer. The schematic diagram of the network is shown in Fig. \ref{fig:NARX}. {To verify that our conclusions are not specific to a particular forecasting model, we additionally perform a robustness analysis based on ensembles generated by XGB models. Since the results are qualitatively similar, they are reported in~\ref{app:sec:xgb}.}

In the NARX model framework, inputs are selected to capture both autoregressive dynamics and the influence of relevant external factors on electricity prices. The choice of inputs is supported by the results of \citep{zie:wer:18}. In the day-ahead electricity market, prices for all 24 hours are established simultaneously one day in advance through an auction \citep{mac:uni:wer:23}. Therefore, the information set available to forecast the price at any hour of the next day is the same. However, since price dynamics are generally different from hour to hour, we treat the prices at each hour of the day as a separate univariate time series and train separate models for each. The first three inputs account for autoregressive effects by including electricity prices for the same hour on days $d-1$, $d-2$, and $d-7$. The price at midnight of the previous day, $p_{d-1,24}$, serves as the last known market value and may signal overnight market behavior. Daily price extremes - $p^{\max}_{d-1}$ and $p^{\min}_{d-1}$ - are included to inform the model of the previous day's price volatility and range. Exogenous inputs also include day-ahead forecasts of total system load and renewable generation, denoted by ${\text{Load}}_{d,h}$ and ${\text{RES}}_{d,h}$, respectively, reflecting expected supply and demand dynamics. {To account for broader market influences, the model incorporates the most recently observed closing prices, from day $d-2$, for key commodities: coal, natural gas, crude oil, and EU carbon emission allowances (EUAs). These variables are included as fundamental drivers of electricity prices because they affect marginal generation costs and the merit-order mechanism, especially during periods of large fuel and EUA price fluctuations such as the European energy crisis \citep{ghe:zie:26}.} In addition, a set of weekday dummies $D_1,\dots,D_7$ captures systematic weekly patterns.

{In this setup we decided to use separate univariate models for each delivery hour, which is a common specification in electricity price forecasting \citep{zie:wer:18}. This choice is motivated by the strong hour-specific structure of day-ahead prices and by the fact that our main objective is not to optimize the point forecasting architecture, but to compare probabilistic postprocessing methods under a controlled and reproducible point-forecasting setup. Similar hourly univariate specifications have been shown to be highly competitive for day-ahead electricity prices \citep{zie:wer:18}.}

The models are retrained daily using a rolling (sliding) window approach, where data from the previous 1456 days (ca. 4 years) are used for model estimation. {At each retraining}, we withdraw a random 10\% of the training data as a validation set for early stopping with a patience of 10 epochs and train the networks using the Levenberg--Marquardt algorithm~\citep{hag:men:1994}. For each day and hour, we generate ensembles of point forecasts from independently trained models. {The ensemble members differ in the randomized initial parameters and the validation split used for early stopping. Unless stated otherwise, the main analysis is based on ensembles containing 25 members. However we provide additional results for ensembles of 10 and 50 members.} Since the differences between the forecasts are primarily caused by the stochastic nature of the training procedure, we treat the ensemble members as exchangeable. Therefore, we sort each ensemble so that the point forecasts used as covariates in the uncertainty quantification methods are non-decreasing in their index, i.e., $\hat{p}_{d,h}^{(1)} \leq \hat{p}_{d,h}^{(2)} \leq \ldots \leq \hat{p}_{d,h}^{(M)}, $
where {$M$ denotes the ensemble size}. We denote the resulting pool of predictions of price $p_{d,h}$ as $\boldsymbol{\hat{p}}_{d,h}$. {The accuracy of all point forecasting model ensembles considered in this study are reported in \ref{app:point_forecasts}.}

\section{Uncertainty quantification methods}
\label{sec:Postprocess}
With an ensemble of point forecasts at our disposal, we can proceed to postprocess them into probabilistic predictions. The general goal is to estimate the probability distribution of the future price $p_{d,h}$ conditional on the price forecasts $\boldsymbol{\hat{p}}_{d, h}$, either in the form of a cumulative distribution function $F_{p_{d,h}}(z|\boldsymbol{\hat{p}}_{d,h})$ or a quantile function $Q_{p_{d,h}}(\tau|\boldsymbol{\hat{p}}_{d,h})$. In our study, we approximate predictive distributions by a set of 99 percentile forecasts, i.e., $\hat{Q}_{p_{d, h}}(\tau|\boldsymbol{\hat{p}}_{d, h})$ for $\tau \in \{\frac{1}{100}, \frac{2}{100}, ..., \frac{99}{100}\}$. 

To provide a comprehensive evaluation of the proposed methods, we compare {iQRA and iSQRA} against a range of state-of-the-art postprocessing techniques for probabilistic electricity price forecasting. We include the original Quantile Regression Averaging (QRA)~\citep{now:wer:15}, {its smoothed variant (SQRA)~\citep{uni:25}}, and the Lasso Quantile Regression Averaging (LQRA)~\citep{uni:wer:21}. Since isotonicity is the key assumption underlying Isotonic Distributional Regression (IDR)~\citep{hen:zie:gne:21}, we also consider IDR as a natural benchmark for the proposed isotonic approaches. Finally, we include Conformal Prediction (CP) and Historical Simulation (HS), two simple yet widely used methods for uncertainty quantification in the machine learning and computational finance literature.

All of the methods we consider are model-agnostic and consistent with the idea of postprocessing -- they work with out-of-sample predictions and can therefore be used without access to the model's training procedure. Estimating the predictive distributions, therefore, requires the set of past forecasts and observations to calibrate the uncertainty quantification models. For each of the methods described below, we use a calibration window of $T = 364$ recent forecasts and re-estimate the models daily. Analogous to the point forecast approach, we compute probabilistic forecasts separately for each hour of the day.

\subsection{Conformal Prediction}
Conformal Prediction (CP) is rapidly gaining attention in various machine learning applications, including probabilistic electricity price forecasting \citep{kat:zie:21, bru:etal:2025}. In regression tasks, it constructs prediction intervals based on out-of-sample prediction errors while maintaining coverage guarantees when the time series are exchangeable \citep{kat:zie:21}. Conformal prediction requires no assumptions about the distribution of prediction errors. On the other hand, in its basic form, only the location of the prediction intervals depends on the point forecast, while their width is influenced only by the empirical distribution of absolute errors in the calibration window.

Despite the fact that prediction intervals derived from CP are valid for any error distribution, translating them into quantile forecasts requires the assumption that the distribution is symmetric. This means that we expect CP to produce a reliable prediction interval, but the mass of errors in the left and right tails are arbitrary. For asymmetric distributions, an analogous method of Historical Simulation (HS) can be applied, which can be thought of as a variant of CP with a conformity score given by non-absolute forecast errors.

\subsection{Isotonic Distributional Regression}
Isotonic Distributional Regression is a novel nonparametric technique that leverages isotonic regression to estimate the CDF of the target variable \citep{hen:zie:gne:21}. The monotonicity constraint, which requires that $\hat{F}(z|x)$ is non-increasing in $x$ for $z\in{\mathbb{R}}$, corresponds to the stochastic order of distributions conditional on the covariate. In the setting of uncertainty quantification, the covariate $x$ is a point prediction from the base regression model (for more details see \citep{lip:uni:wer:24}). In essence, the isotonicity of the distribution means that greater point predictions imply a stochastically bigger target variable. For a fixed $z$, $\hat{F}(z|x)$ is estimated as a solution to the isotonic least squares problem, which corresponds to minimizing the continuous ranked probability score \citep{hen:zie:gne:21} -- a strictly consistent scoring function for probability distributions.

Since we use an ensemble of {$M = \{10, 25, 50\}$} point forecasts as input to the uncertainty quantification, we need to choose an approach to estimate distribution functions conditional on this set of regressors. We tested several approaches: ordering the ensembles with the component-wise order (which in our case corresponds to the empirical stochastic order, since our covariates are sorted) or the increasing convex order \citep{hen:zie:gne:21}; estimating a single IDR with the ensemble mean as the regressor (analogous to the committee machine setting of quantile regression averaging); and a linear pool of independently estimated IDRs, one for each point forecast in the ensemble. We present results for the latter approach because it outperformed the former in {preliminary testing}.

\subsection{Quantile Regression Averaging}
Quantile regression is a general method for estimating the quantiles of target variables as linear functions of covariates. \citet{tay:bun:1998} proposed using quantile regression to combine different quantile estimates, later \citet{now:wer:15} introduced quantile regression averaging, which combines a pool of point forecasts to predict a target quantile. Since then, it has been widely used in various energy forecasting tasks 
\citep{liu:now:hon:wer:17,yan:yan:liu:23,cor:din:pou:24}, achieving top results in the GEFCom 2014 competition \citep{mac:now:16, gai:gou:ned:16} 
and subsequently establishing itself as a method for quantifying uncertainty in electricity price forecasting \citep{mac:uni:wer:23}. We adapt the multiple quantile regression approach \citep{bro:gil:2020}, where we estimate a separate model for each quantile. The quantile forecasts resulting from independently estimated models may {fail to be non-decreasing}; in which case, we sort the forecasts to obtain a consistent set of quantile forecasts. The model for each of the 99 percentiles is of the form:
\begin{equation}
\label{eq:lin:quantiles}
  \hat{Q}_{p_{d, h}}(\tau) = \beta_0 + \boldsymbol{\beta}^T \boldsymbol{\hat{p}}_{d, h}
\end{equation}
with the coefficients $\beta_0$ and $\boldsymbol{\beta}$ estimated by solving a linear programming problem of minimizing the pinball loss for a corresponding probability $\tau$ on a calibration window of $T$ time steps:
\begin{equation}
\label{eq:qrmin}
  \underset{\beta_0, \boldsymbol{\beta}}{\argmin} \sum_{t=d-T}^{d-1}(\mathbf{1}_{[p_{t, h} < \hat{Q}_{p_{t, h}}(\tau)]} - \tau) (\hat{Q}_{p_{t, h}}(\tau)- p_{t, h})
\end{equation}

In addition to the described QRA approach, we also include its committee machine version, called Quantile Regression Machine (QRM)~\citep{mar:uni:wer:2020}, which corresponds to calculating the average of the ensemble of forecasts and treating it as a single regressor in quantile regression. {Incorporating QRM into the analysis allows us to answer whether using the full ensemble can provide additional predictive power, or if the first moment alone is sufficient.}

{Note that QRM is also closely related to Factor QRA (FQRA), since the single-factor version of FQRA reduces the ensemble to a similar signal: the average level of the forecasts \citep{mac:ser:uni:24}. Since FQRA was found to perform worse than QRM in that setting, we do not include it as a separate benchmark.}

\subsection{Smoothed Quantile Regression Averaging}
{Recently, \citet{fer:gue:hor:21} proposed smoothed quantile regression (SQR) as an alternative to standard quantile regression. The key idea is to replace the non-differentiable pinball loss with a smoothing approximation obtained using kernel estimation. The resulting estimator enjoys favorable statistical properties and has been shown to reduce estimation error relative to standard quantile regression.

Following \citet{uni:25}, we apply smoothed quantile regression to the forecast combination problem. Analogously to QRA, a separate model is estimated for each quantile and the target quantile is expressed as a linear combination of the ensemble forecasts. However, instead of minimizing the pinball loss, the model parameters are obtained by solving

\begin{equation}
\underset{\beta_0,\boldsymbol{\beta}}{\argmin}
\sum_{t=d-T}^{d-1}
\left[
H\phi\left(\frac{p_{t,h}-\hat{Q}_{p_{t,h}}(\tau)}{H}\right)
+\left(\tau-\Phi\left(\frac{\hat{Q}_{p_{t,h}}(\tau) - p_{t,h}}{H}\right)\right)
\left(p_{t,h}-\hat{Q}_{p_{t,h}}(\tau)\right)
\right],
\end{equation}

where $\Phi(\cdot)$ and $\phi(\cdot)$ denote the cumulative distribution function and probability density function of the standard normal distribution, respectively, and $H$ is the bandwidth parameter controlling the degree of smoothing. Following \citet{uni:25}, the bandwidth is selected using \citep{sco:92} rule-of-thumb. As in the case of QRA, independently estimated quantiles are sorted whenever necessary to ensure a non-decreasing quantile function.}

\subsection{Lasso Quantile Regression Averaging}
Since our ensemble consists of a relatively large number of forecasts, we consider the quantile regression problem with the Lasso penalty for the purpose of variable selection \citep{tib:1996}, which has been shown to outperform the unregularized QRA in probabilistic electricity price forecasting based on the same {ensemble size of $M = \{10, 25, 50\}$ input} forecasts \citep{uni:wer:21}. Quantile forecasts are parametrized by the same linear formula given by Eq.\eqref{eq:lin:quantiles}, but the optimal coefficients minimize the regularized loss function:

\begin{equation}
  \underset{\beta_0, \boldsymbol{\beta}}{\argmin} \sum_{t=d-T}^{d-1}(\mathbf{1}_{[p_{t, h} < \hat{Q}_{p_{t, h}}(\tau)]} - \tau) (\hat{Q}_{p_{t, h}}(\tau) - p_{t, h}) + \lambda \left\| \boldsymbol{\beta}\right\|_1
\end{equation}

where $\lambda$ is the strength of the regularization. Following \citet{uni:wer:21} we select the optimal $\lambda$ for each percentile separately. We train 20 models with different $\lambda$ values, with the regularization path ranging from $\lambda_{\max}$ to $10^{-5}\lambda_{max}$ on a log scale grid and then select the best model according to the Bayesian Information Criterion (BIC) computed on the calibration window. The upper limit of the regularization path is analytically derived as the largest feasible regularization strength~\citep{li:zhu:2008}.

\subsection{Isotonic {(Smoothed)} Quantile Regression Averaging}
\label{ssec:iqra}

The isotonic constraint employed in the IDR regularizes the solution of the CDF estimation problem. The stochastic order described by the monotonicity of the cumulative distribution function $F(z|x)$ in $x$ can be equivalently described by the monotonicity of the quantile function $Q(\tau|x)$. A conditional distribution that is isotonic in $x$ is described by $Q(\tau|x)$ that is monotonically nondecreasing in $x$. In fact, the idea of estimating isotonic quantile functions precedes the IDR and was already considered by \citet{cas:cry:1976}, who proposed a min-max estimator for nonparametric isotonic quantile functions. The equivalence between this min-max optimization and pinball loss minimization under isotonicity constraints was proved in \citep{poi:tho:2000}. The nonparametric isotonic quantiles correspond to the distribution functions estimated by IDR, see \citep{mos:dum:2020} for details on their convergence.

The concept of isotonic quantiles is far from new, but to the best of our knowledge, isotonicity constraints have not been considered in {linear forecast combination methods for probabilistic forecasting}. The linear form of {the quantile functions in both QRA and SQRA} allows us to easily impose isotonicity {by constraining the coefficients (except for the intercept) to be nonnegative}. {Consequently, we introduce two isotonic variants of the benchmark methods considered in this study: Isotonic Quantile Regression Averaging (iQRA) and Isotonic Smoothed Quantile Regression Averaging (iSQRA).}

{For iQRA, the coefficients are obtained by solving the quantile regression problem in Eq.~\eqref{eq:qrmin} subject to} ${
\beta_i \geq 0, \; i=1,\ldots,M,
}$
{where $M$ denotes the ensemble size. Analogously, iSQRA is obtained by solving the smoothed quantile regression problem subject to the same isotonicity constraints. Since the ensemble forecasts are sorted in ascending order prior to model estimation, nonnegative coefficients guarantee that the estimated quantiles are monotonically nondecreasing with respect to each ensemble member. Moreover, introducing isotonic regularization through constrained coefficients reduces the complexity of the linear program underlying quantile regression, leading to a reduction in computational cost. We discuss this further from the perspective of both practical runtimes and theoretical asymptotic complexity in Section~\ref{sec:Results}. The proposed method is an interpretable, more efficient and scalable alternative to variable selection with lasso, without compromising the predictive performance.}

The constraint can be easily implemented in the linear programming formulation of quantile regression by reducing the search space. For example, expressing quantile regression as a linear program in from $\min_{Ax=b, x\geq 0}c^\text{T}x$ requires decomposing each coefficient (including the intercept) into a negative and positive part, ${\beta_i}{\in\mathbb{R}} = {\beta_i^+}{\in\mathbb{R^+}} - {\beta_i^-}{\in\mathbb{R^+}}$. Thus, isotonicity can be enforced by simply eliminating the $\beta_i^-$ variables from the linear program. Therefore, introducing a stochastic order constraint reduces the complexity of {the estimation procedure} without the need for additional penalty terms or hyperparameters. Noteworthy, isotonicity imposed by positive weights was previously considered by \citet{can:2018} in mononote quantile regression neural networks.

In the literature, the term \textit{monotone quantiles} is often used in relation to the isotonic property of the quantile function \citep{bol:eil:aer:2006, tak:etal:2006, mug:sci:aug:2012, can:2018}. However, monotonicity can also refer to the probability value of the quantile. This is always required by a proper quantile function, but methods such as multiple quantile regression can produce nonmonotonic quantile estimates, a problem widely known as \textit{quantile crossing}. {Therefore, we call the method isotonic quantile regression, invoking the association with IDR, to highlight that we focus on the isotonic relation of quantiles and covariates, not the quantile crossing problem.}

\section{Results}
\label{sec:Results}
\subsection{Validation measure}
As emphasized by Gneiting and Raftery \citep{gne:raf:07}, the evaluation of probabilistic predictions should aim at maximizing the sharpness subject to reliability. Reliability refers to the extent to which the prediction intervals contain the observed values, i.e. the empirical coverage. Sharpness, on the other hand, measures the concentration of the predictive distribution, which is typically reflected in the width of the prediction intervals. Ideally, the intervals should be as narrow as possible while maintaining the desired coverage level. To assess both reliability and sharpness, we use a range of evaluation metrics that together reflect the quality of our probabilistic predictions.

First, to test the reliability we propose to use the Average Coverage Error (ACE) of the prediction intervals, i.e., the difference between the fraction of observations that fell inside the prediction interval (empirical coverage) and its nominal coverage \citep{now:wer:18} {over a test set $\mathcal{D}$}:
\begin{eqnarray}
\label{eq:ace}
  {\text{ACE}(\alpha) = \alpha - \left(\frac{1}{24 |\mathcal{D}|}\sum_{d\in \mathcal{D}}\sum_{h=1}^{24}  \mathbf{1}_{\{p_{d,h} \in [\hat{L}_{d, h}, \hat{U}_{d, h}]\}}\right)},
\end{eqnarray}
where the bounds of a $\alpha$-PI are derived from our percentile forecasts according to $\hat{L}_{d, h} = \hat{Q}_{p_{d, h}}(\frac{1-\alpha}{2})$ and $\hat{U}_{d, h} = \hat{Q}_{p_{d, h}}(\frac{1+\alpha}{2})$. 

A convenient way to jointly assess both reliability and sharpness is through the Pinball Score (PS), a proper scoring rule \citep{pet:etal:22} commonly used in the electricity price forecasting (EPF) literature \citep{now:wer:18}. The measure is defined as:
\begin{equation}\label{eq:Pinball}
	{\mbox{PS}_{d,h} (\tau) = 
  \left(\mathbf{1}_{P_{d,h} < \hat{P_{d,h}^\tau}} - \tau \right) \left(\hat{P}^{\tau}_{d,h} - P_{d,h}\right)}
\end{equation}
where $\hat{P}^{\tau}_{d,h}$ is the forecast of the price quantile of order ${\tau} \in(0,1)$ and $P_{d,h}$ is the observed price for day $d$ and hour $h$.

In this paper, we use the pinball score to assess the quality of the interval forecasts and define a Prediction Interval Pinball Score (PIPS):
\begin{equation}
\label{eq:pips}
  \text{PIPS}_{d, h}(\alpha) = \tfrac{1}{2}\mbox{PS}_{d,h} \left(\tfrac{1-\alpha}{2}\right) + \tfrac{1}{2}\mbox{PS}_{d,h}\left(\tfrac{1+\alpha}{2}\right),
\end{equation}
which is a proper scoring rule for prediction intervals constructed from the {$\frac{1-\alpha}{2}$ and $\frac{1+\alpha}{2}$} quantile forecasts \citep{gne:raf:07}.

Finally, to take into account not only selected prediction intervals but the entire predictive distribution, we use the continuous ranked probability score (CRPS) \citep{gne:raf:07}. The CRPS is a proper scoring rule and the standard metric for evaluating probabilistic electricity price forecasts \citep{mac:uni:wer:23}. It is defined as
\begin{equation}
  \operatorname{CRPS}(\hat{F}, x) = \int_{-\infty}^{\infty} \left(\hat{F}(y) - \mathbf{1}_{\lbrace x \leq y \rbrace} \right)^2 dy\approx 2 \int_0^1 \text{PS}(\tau)\,d\tau,
\end{equation}
where $\hat{F}$ is the predictive distribution and $x$ is the observation, for example, the electricity price $P_{d,h}$. It can be approximated by:\footnote{Note that {formally the scaling factor of 2 should be included in Eq. \eqref{eqn:CRPS:approx}, but it is} usually omitted in practice \citep{nit:wer:23}. This is also the case here.}
{
\begin{equation}\label{eqn:CRPS:approx}
  \operatorname{CRPS}_{d,h}(\hat{F}, x) \approx {\frac{1}{K}} \sum_{i=1}^K \operatorname{PS}_{d,h}\left(\tau_i\right),
\end{equation}
where $\left(\tau_1, \ldots, \tau_K\right)$} is an equidistant dense grid of probabilities, e.g. the 99 percentiles.

\subsection{Empirical results}

\begin{table}[b!]
\centering
\caption{{Continuous ranked probability score (CRPS) for a postprocessed ensemble of 25 NARX point forecasts and {six} years of out-of-sample period on German and French markets}.}
\label{tab:CRPS}
\scalebox{1}{
\begin{tabular}{c|c|c|c|c|c|c|c|c|c|c|c|c}
\multicolumn{1}{l}{}  & \multicolumn{6}{c|}{\textbf{Germany}} & \multicolumn{6}{c}{\textbf{France}}  \\
\multicolumn{1}{l|}{\textbf{}} & \textbf{2020}  & \textbf{2021}  & \textbf{2022} & \textbf{2023}  & \textbf{2024}  & \textbf{2025}  & \textbf{2020}  & \textbf{2021}  & \textbf{2022} & \textbf{2023}  & \textbf{2024}  & \textbf{2025}  \\
\textbf{HS} & \cellcolor[HTML]{F6FBF8}1.53 & \cellcolor[HTML]{FFF3B3}4.71 & \cellcolor[HTML]{FFFEF4}10.90 & \cellcolor[HTML]{FFF09F}5.79 & \cellcolor[HTML]{FFFCE9}5.12 & \cellcolor[HTML]{FFF1A6}5.23 & \cellcolor[HTML]{F3FAF5}1.46 & \cellcolor[HTML]{FFF6C3}5.32 & \cellcolor[HTML]{FFFDF0}12.49 & \cellcolor[HTML]{FFF3B0}5.99 & \cellcolor[HTML]{63BE7B}5.14 & \cellcolor[HTML]{FFFEF5}5.04 \\
\textbf{CP} & \cellcolor[HTML]{99D4A9}1.51 & \cellcolor[HTML]{FFF2AA}4.73 & \cellcolor[HTML]{FFFEF5}10.89 & \cellcolor[HTML]{FFEF9C}5.80 & \cellcolor[HTML]{FFFEF5}5.10 & \cellcolor[HTML]{FFEF9C}5.26 & \cellcolor[HTML]{CBE9D3}1.45 & \cellcolor[HTML]{FFEF9C}5.43 & \cellcolor[HTML]{FFFEF4}12.46 & \cellcolor[HTML]{FFF6C4}5.92 & \cellcolor[HTML]{FFFFFF}5.18 & \cellcolor[HTML]{FFFADB}5.13 \\
\textbf{IDR}  & \cellcolor[HTML]{FFEF9C}1.60 & \cellcolor[HTML]{FFEF9C}4.76 & \cellcolor[HTML]{FFF5BE}11.19 & \cellcolor[HTML]{63BE7B}5.11 & \cellcolor[HTML]{FFFFFE}5.08 & \cellcolor[HTML]{63BE7B}4.69 & \cellcolor[HTML]{FFEF9C}1.51 & \cellcolor[HTML]{FFF3B0}5.38 & \cellcolor[HTML]{FFF8D3}12.71 & \cellcolor[HTML]{63BE7B}5.53 & \cellcolor[HTML]{F8FCF9}5.17 & \cellcolor[HTML]{63BE7B}4.83 \\
\textbf{QRA}  & \cellcolor[HTML]{FFF09E}1.60 & \cellcolor[HTML]{FFF4BA}4.69 & \cellcolor[HTML]{FFEF9C}11.38 & \cellcolor[HTML]{FFF5BD}5.69 & \cellcolor[HTML]{FFEF9C}5.25 & \cellcolor[HTML]{FFF4B7}5.16 & \cellcolor[HTML]{FFF3B2}1.50 & \cellcolor[HTML]{FFF3B0}5.38 & \cellcolor[HTML]{FFEF9C}13.14 & \cellcolor[HTML]{FFEF9C}6.04 & \cellcolor[HTML]{FFEF9C}5.43 & \cellcolor[HTML]{FFEF9C}5.33 \\
\textbf{SQRA} & \cellcolor[HTML]{FFF1A8}1.59 & \cellcolor[HTML]{FFF4B7}4.70 & \cellcolor[HTML]{FFF1A8}11.32 & \cellcolor[HTML]{FFF5BC}5.70 & \cellcolor[HTML]{FFF2AA}5.23 & \cellcolor[HTML]{FFF5BD}5.14 & \cellcolor[HTML]{FFF6C2}1.49 & \cellcolor[HTML]{FFF2AC}5.39 & \cellcolor[HTML]{FFF1A6}13.07 & \cellcolor[HTML]{FFF09E}6.04 & \cellcolor[HTML]{FFF2AF}5.39 & \cellcolor[HTML]{FFF1A7}5.30 \\
\textbf{QRM}  & \cellcolor[HTML]{FFFEF9}1.53 & \cellcolor[HTML]{B3DFBF}4.41 & \cellcolor[HTML]{D0EBD8}10.77 & \cellcolor[HTML]{F8FCF9}5.46 & \cellcolor[HTML]{CDEAD5}5.06 & \cellcolor[HTML]{FDFEFD}4.89 & \cellcolor[HTML]{FFFEF7}1.47 & \cellcolor[HTML]{A5D9B2}4.98 & \cellcolor[HTML]{B2DFBE}12.28 & \cellcolor[HTML]{FAFCFA}5.74 & \cellcolor[HTML]{A9DBB6}5.15 & \cellcolor[HTML]{FFFFFF}5.01 \\
\textbf{SQRM} & \cellcolor[HTML]{FFFFFD}1.53 & \cellcolor[HTML]{AADBB7}4.39 & \cellcolor[HTML]{B3DFBE}10.74 & \cellcolor[HTML]{FFFFFA}5.49 & \cellcolor[HTML]{C1E5CA}5.05 & \cellcolor[HTML]{FFFFFF}4.90 & \cellcolor[HTML]{FFFEF9}1.47 & \cellcolor[HTML]{9CD6AB}4.96 & \cellcolor[HTML]{98D4A8}12.25 & \cellcolor[HTML]{FFFFFD}5.75 & \cellcolor[HTML]{80CA93}5.14 & \cellcolor[HTML]{FFFFFF}5.01 \\
\textbf{LQRA} & \cellcolor[HTML]{AFDDBB}1.51 & \cellcolor[HTML]{6AC181}4.27 & \cellcolor[HTML]{63BE7B}10.64 & \cellcolor[HTML]{B1DEBD}5.30 & \cellcolor[HTML]{F8FCF9}5.08 & \cellcolor[HTML]{BBE2C5}4.81 & \cellcolor[HTML]{84CB97}1.43 & \cellcolor[HTML]{64BE7B}4.85 & \cellcolor[HTML]{88CD9A}12.23 & \cellcolor[HTML]{A8DAB5}5.63 & \cellcolor[HTML]{FFFBE1}5.25 & \cellcolor[HTML]{E7F5EB}4.98 \\
\textbf{iQRA} & \cellcolor[HTML]{6DC284}1.50 & \cellcolor[HTML]{68C07F}4.26 & \cellcolor[HTML]{94D2A4}10.70 & \cellcolor[HTML]{C9E8D1}5.35 & \cellcolor[HTML]{79C78E}5.02 & \cellcolor[HTML]{D4EDDA}4.84 & \cellcolor[HTML]{6AC181}1.43 & \cellcolor[HTML]{69C080}4.86 & \cellcolor[HTML]{83CB96}12.22 & \cellcolor[HTML]{CFEBD6}5.68 & \cellcolor[HTML]{FFFFFE}5.18 & \cellcolor[HTML]{F3FAF5}5.00 \\
\textbf{iSQRA}  & \cellcolor[HTML]{63BE7B}1.50 & \cellcolor[HTML]{63BE7B}4.25 & \cellcolor[HTML]{71C487}10.65 & \cellcolor[HTML]{D3ECDA}5.38 & \cellcolor[HTML]{63BE7B}5.01 & \cellcolor[HTML]{D6EEDC}4.84 & \cellcolor[HTML]{63BE7B}1.42 & \cellcolor[HTML]{63BE7B}4.85 & \cellcolor[HTML]{63BE7B}12.19 & \cellcolor[HTML]{D5EDDC}5.69 & \cellcolor[HTML]{D3ECDA}5.16 & \cellcolor[HTML]{F1F9F3}4.99
\end{tabular}
}
\end{table}

{Table~\ref{tab:CRPS} reports the continuous ranked probability scores (CRPS) for the German and French markets over the six-year out-of-sample period. Overall, the proposed isotonic approaches remain highly competitive, with iQRA and iSQRA consistently ranking among the top-performing methods. Their performance is comparable to that of LQRA, with only small differences in CRPS observed across years and markets. 

{The year-by-year results also illustrate the changing difficulty of the forecasting task. Forecast errors increase markedly in 2022, reflecting the extreme price dynamics during the European energy crisis, whereas the pre-crisis and post-crisis years exhibit substantially lower CRPS values. Importantly, despite these changes in market conditions, LQRA, iQRA, and iSQRA remain the leading group of methods in most years, suggesting that the relative performance of the regularized postprocessing approaches is stable across markedly different regimes.}

Among the remaining methods, IDR achieves competitive results in 2023 and 2025, while QRM and SQRM generally outperform the corresponding QRA and SQRA models. In contrast, the unregularized QRA and SQRA methods typically yield the highest CRPS values among the quantile regression approaches, confirming the benefits of isotonic regularization for probabilistic electricity price forecasting.}

\begin{figure}[tb!]
  \centering
  \begin{minipage}{0.49\textwidth}
  \centering  
  \textbf{\quad Germany}
  
  \includegraphics[width=\linewidth]{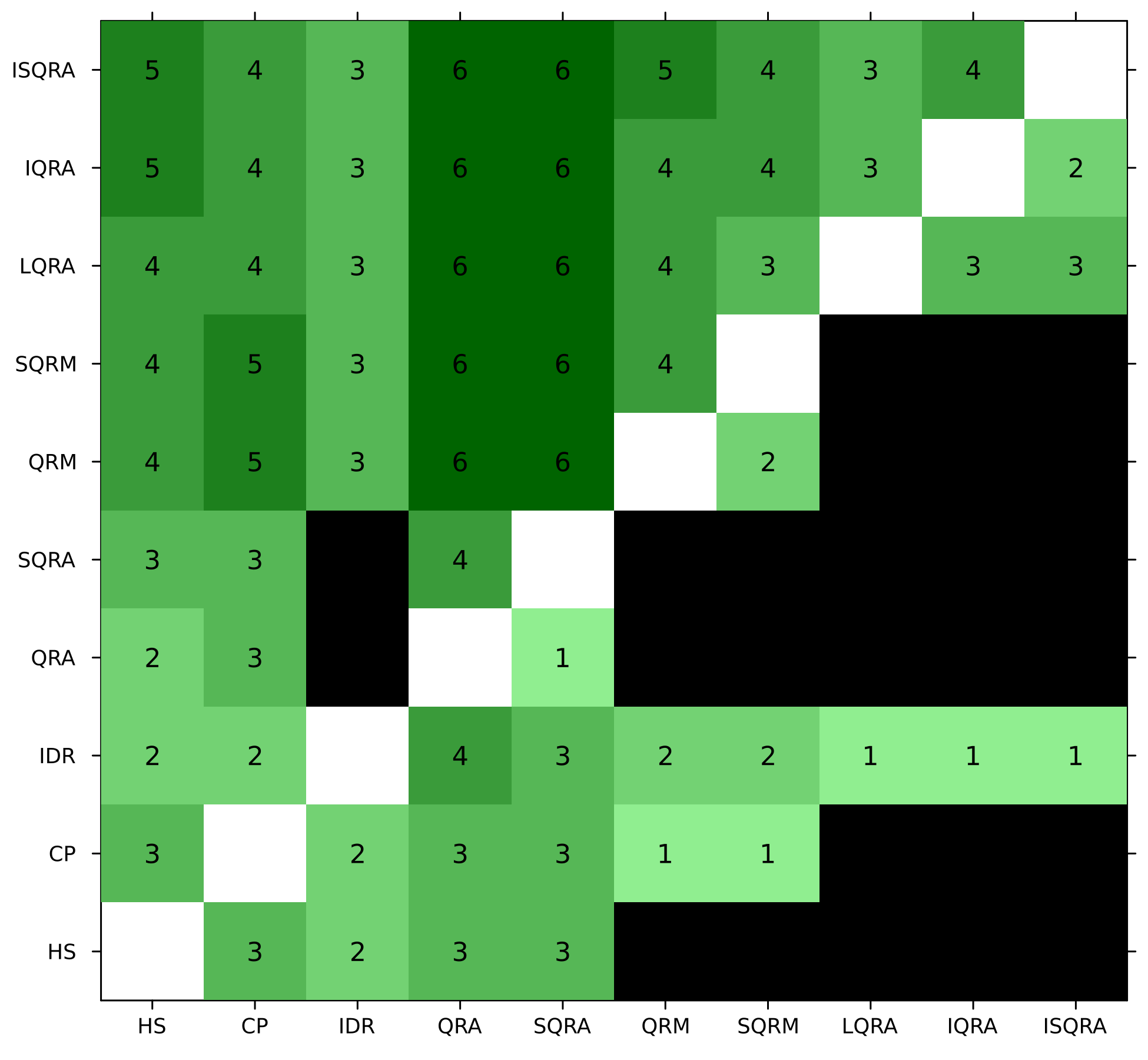}
  \end{minipage}
  \begin{minipage}{0.49\textwidth}
  \centering  
  \textbf{\quad France}
  
  \includegraphics[width=\linewidth]{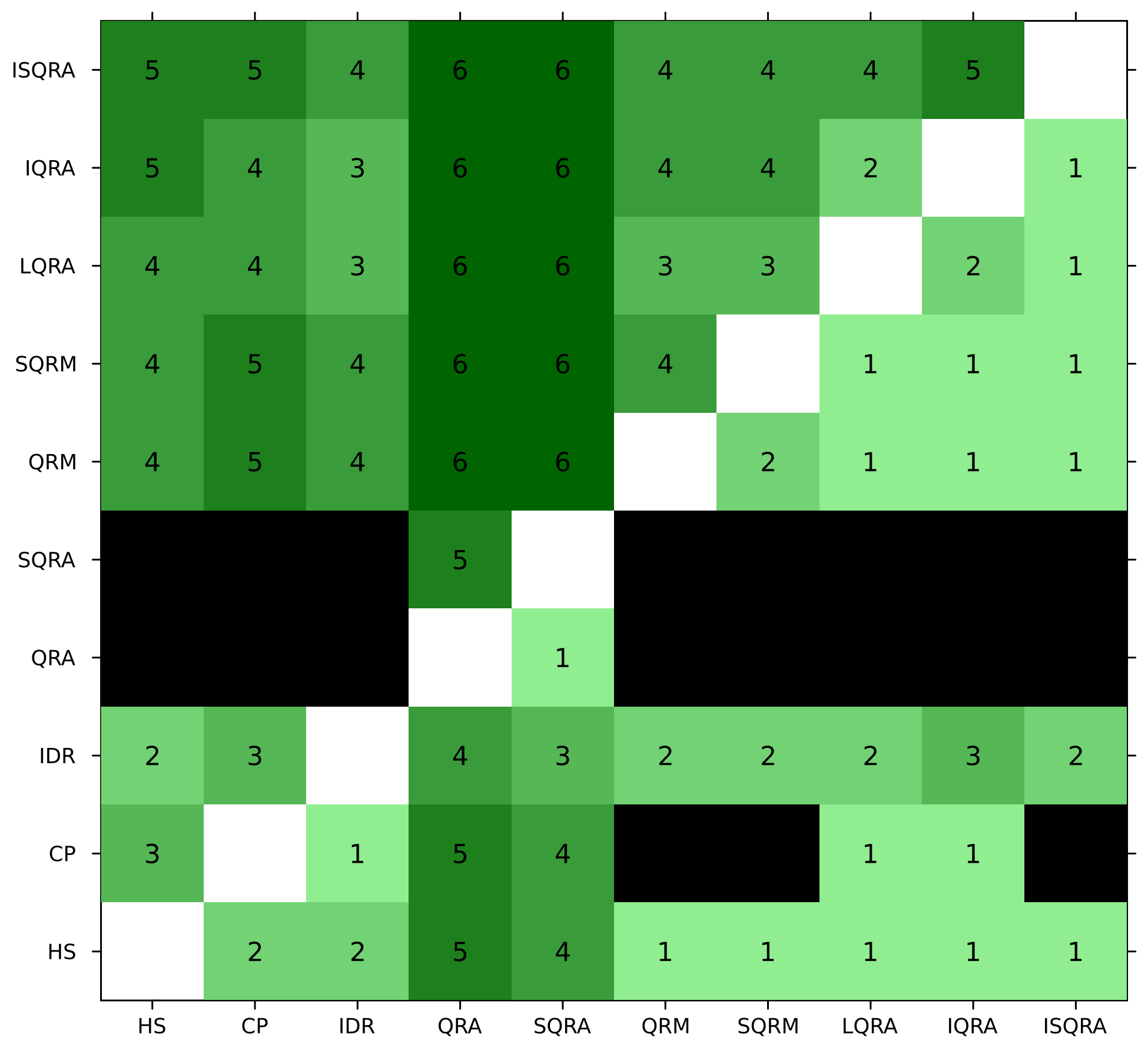}
  \end{minipage}
  \caption{Number of years in which the model on the vertical axis outperformed the model on the horizontal axis according to the Giacomini-White test at the 1\% significance level. Results obtained for an ensemble of 25 NARX forecasts.}
  \label{fig:gw}
\end{figure}

{Following \citep{lip:uni:wer:24} we employ the test of \textit{conditional predictive ability} \citep[CPA;][]{gia:whi:06} to formally compare the predictive performance of the competing probabilistic forecasting methods. Unlike the standard application of the CPA test, we perform the test separately for each calendar year using the daily CRPS values and fix the significance level at $1\%$.
The results for all pairs of models and both markets are summarized in Figure~\ref{fig:gw}. Each cell reports the number of years (out of six) in which the model on the y-axis significantly outperformed the model on the x-axis according to the CPA test. Consequently, larger values indicate a more consistent predictive advantage of the model on the vertical axis, whereas black cells (value 0) indicate that no statistically significant superiority was detected in any year.}

{The CPA results largely confirm the conclusions drawn from Table~\ref{tab:CRPS}. In particular, the proposed isotonic methods, iQRA and iSQRA, consistently outperform their non-isotonic counterparts, QRA and SQRA, with statistically significant improvements observed in most years and for both markets. Likewise, the CPA test confirms that LQRA, iQRA, and iSQRA constitute the leading group of methods. While their CRPS values are very similar, none of the three methods consistently dominates the other two across all years, indicating comparable predictive performance. Finally, the similarity of the CPA matrices for the German and French markets demonstrates that these conclusions remain robust under different market conditions.}

{To investigate the effect of the ensemble size on probabilistic forecasting performance, we repeat the analysis using ensembles of 10, 25, and 50 independently trained NARX models. The results, aggregated over the entire out-of-sample period, are reported in Table~\ref{tab:CRPS:ens}.}

{Several interesting observations can be made from Table~\ref{tab:CRPS:ens}. First, increasing the ensemble size does not necessarily improve probabilistic forecasts. In fact, the performance of QRA and SQRA deteriorates considerably when the number of ensemble members increases from 25 to 50. In contrast, the remaining methods are much less sensitive to the ensemble size. In particular, LQRA, iQRA, and iSQRA exhibit very stable performance and even achieve slightly lower CRPS values for larger ensembles. As in the previous experiments, the three methods remain the best-performing approaches across both markets, with only marginal differences between them.}

\begin{table}[tbh!]
\centering
\caption{{Continuous ranked probability score (CRPS) for different sizes of the ensemble of NARX point forecasts, on German and French markets, aggregated for the entire out-of-sample testing period.}}
\label{tab:CRPS:ens}
\scalebox{1}{
\begin{tabular}{c|c|c|c|c|c|c}
\multicolumn{1}{l}{}  & \multicolumn{3}{c|}{\textbf{Germany}} & \multicolumn{3}{c}{\textbf{France}}  \\
\multicolumn{1}{l|}{\textbf{}} & \textbf{10}  & \textbf{25}  & \textbf{50}  & \textbf{10}  & \textbf{25}  & \textbf{50}  \\
\textbf{HS} & \cellcolor[HTML]{FFFADF}5.592 & \cellcolor[HTML]{FFFCE7}5.542 & \cellcolor[HTML]{FFFCEA}5.525 & \cellcolor[HTML]{FFFDEF}5.960 & \cellcolor[HTML]{FFFEF7}5.904 & \cellcolor[HTML]{FFFEF8}5.893 \\
\textbf{CP} & \cellcolor[HTML]{FFFADE}5.594 & \cellcolor[HTML]{FFFCE7}5.544 & \cellcolor[HTML]{FFFCEA}5.527 & \cellcolor[HTML]{FFFCEC}5.982 & \cellcolor[HTML]{FFFEF4}5.926 & \cellcolor[HTML]{FFFEF5}5.915 \\
\textbf{IDR}  & \cellcolor[HTML]{FFFFFB}5.428 & \cellcolor[HTML]{FFFFFF}5.403 & \cellcolor[HTML]{FFFFFF}5.404 & \cellcolor[HTML]{FFFFFA}5.880 & \cellcolor[HTML]{FFFFFE}5.853 & \cellcolor[HTML]{FFFFFD}5.856 \\
\textbf{QRA}  & \cellcolor[HTML]{FFFEF6}5.455 & \cellcolor[HTML]{FFF9D9}5.626 & \cellcolor[HTML]{FFEF9C}5.973 & \cellcolor[HTML]{FFFEF4}5.922 & \cellcolor[HTML]{FFF9D8}6.134 & \cellcolor[HTML]{FFEF9C}6.578 \\
\textbf{SQRA} & \cellcolor[HTML]{FFFEF9}5.440 & \cellcolor[HTML]{FFFADC}5.609 & \cellcolor[HTML]{FFF09E}5.965 & \cellcolor[HTML]{FFFEF7}5.903 & \cellcolor[HTML]{FFFADB}6.109 & \cellcolor[HTML]{FFF09D}6.576 \\
\textbf{QRM}  & \cellcolor[HTML]{FDFEFD}5.400 & \cellcolor[HTML]{CFEBD6}5.352 & \cellcolor[HTML]{BFE4C8}5.335 & \cellcolor[HTML]{EFF8F1}5.825 & \cellcolor[HTML]{B2DEBD}5.769 & \cellcolor[HTML]{A3D9B1}5.756 \\
\textbf{SQRM} & \cellcolor[HTML]{F9FCFA}5.396 & \cellcolor[HTML]{CBE9D3}5.348 & \cellcolor[HTML]{BBE2C5}5.331 & \cellcolor[HTML]{E7F5EB}5.818 & \cellcolor[HTML]{A9DBB6}5.761 & \cellcolor[HTML]{9BD5AA}5.748 \\
\textbf{LQRA} & \cellcolor[HTML]{C0E4C9}5.336 & \cellcolor[HTML]{7BC88F}5.264 & \cellcolor[HTML]{63BE7B}5.239 & \cellcolor[HTML]{C1E5CA}5.783 & \cellcolor[HTML]{83CB96}5.727 & \cellcolor[HTML]{74C589}5.713 \\
\textbf{iQRA} & \cellcolor[HTML]{C5E7CE}5.342 & \cellcolor[HTML]{86CC99}5.276 & \cellcolor[HTML]{71C387}5.254 & \cellcolor[HTML]{BBE3C6}5.778 & \cellcolor[HTML]{81CA95}5.725 & \cellcolor[HTML]{6DC284}5.707 \\
\textbf{iSQRA}  & \cellcolor[HTML]{C0E4C9}5.336 & \cellcolor[HTML]{7FCA93}5.269 & \cellcolor[HTML]{6AC181}5.247 & \cellcolor[HTML]{B2DEBD}5.769 & \cellcolor[HTML]{76C68B}5.715 & \cellcolor[HTML]{63BE7B}5.697
\end{tabular}
}

\end{table}

{Finally, to provide a statistically rigorous comparison of the models, we perform the Model Confidence Set test~\citep{han:etal:11} over the entire out-of-sample period, separately for each market and ensemble size. The results, presented in Table~\ref{tab:MCS}, show that LQRA, iQRA and iSQRA are retained in the 99\% confidence set for all markets and ensemble sizes. Except for ensembles of 10 networks, these are the only models retained in this set. LQRA and iSQRA are also retained in the 95\% confidence set in all cases, whereas iQRA is excluded only for ensembles of 25 and 50 networks in the German market.}

\begin{table}[tb!]
\centering

\caption{{The results of the Model Confidence Set based on the daily CRPS values and the $T_{\text{max}}$ statistic~\citep{han:etal:11}. The columns correspond to two markets and three different ensemble sizes. The test was run with 10,000 bootstrap replications and two significance levels: $\alpha = 1\%$ and $\alpha = 5\%$. One checkmark indicates that the model was included in the 99\% confidence set, two checkmarks denote models included in the 95\% confidence set. Models omitted from the table were {included in the test, but} not retained in either confidence set.}}
\label{tab:MCS}
\scalebox{1}{
\begin{tabular}{c|c|c|c|c|c|c}
\multicolumn{1}{l}{}  & \multicolumn{3}{c|}{\textbf{Germany}}  & \multicolumn{3}{c}{\textbf{France}} \\
\multicolumn{1}{l|}{\textbf{}} & \textbf{10} & \textbf{25} & \textbf{50}  & \textbf{10}  & \textbf{25} & \textbf{50} \\
\textbf{IDR} & \checkmark  & &  & \checkmark & & \\
\textbf{QRM} & & &  & \checkmark & & \\
\textbf{SQRM} & & &  & \checkmark & & \\
\textbf{LQRA} & \checkmark \checkmark  & \checkmark \checkmark  & \checkmark \checkmark & \checkmark \checkmark & \checkmark \checkmark & \checkmark \checkmark \\
\textbf{iQRA}  & \checkmark \checkmark & \checkmark & \checkmark & \checkmark \checkmark & \checkmark \checkmark & \checkmark \checkmark \\
\textbf{iSQRA} & \checkmark \checkmark & \checkmark \checkmark & \checkmark \checkmark & \checkmark \checkmark & \checkmark \checkmark & \checkmark \checkmark \\
\end{tabular}
}
\end{table}

{Note that the conclusions for the XGB-based prediction are very {similar}. On average the result presented in \ref{app:sec:xgb}} are slightly worse, but the key finding stays the same.

\subsubsection{Coverage and tail performance}

\begin{table}[t!]
\centering
\caption{{Average Coverage Error (\%) for the ensemble of 25 NARX point forecasts and four prediction interval levels, across two markets.}}
\label{tab:ACE}
\scalebox{1}{
\begin{tabular}{c|c|c|c|c|c|c|c|c}
\multicolumn{1}{l}{}  & \multicolumn{4}{c|}{\textbf{Germany}}  & \multicolumn{4}{c}{\textbf{France}} \\
\multicolumn{1}{l|}{\textbf{}} & \textbf{1-99\%} & \textbf{2-98\%} & \textbf{5-95\%}  & \textbf{10-90\%}  & \textbf{1-99\%} & \textbf{2-98\%} & \textbf{5-95\%}  & \textbf{10-90\%}  \\
\textbf{HS} & \cellcolor[HTML]{FCFEFD}1.99 & \cellcolor[HTML]{FFFFFC}2.68 & \cellcolor[HTML]{FFFCEA}4.21 & \cellcolor[HTML]{FFFBE1}5.08 & \cellcolor[HTML]{FFFFFB}2.08 & \cellcolor[HTML]{FFFDF3}2.84 & \cellcolor[HTML]{FFFCEC}3.90 & \cellcolor[HTML]{FFFBE4}4.79 \\
\textbf{CP} & \cellcolor[HTML]{7AC78E}1.44 & \cellcolor[HTML]{C5E7CE}2.18 & \cellcolor[HTML]{FFFEF8}3.58 & \cellcolor[HTML]{FFFDF0}4.57 & \cellcolor[HTML]{CAE9D2}1.58 & \cellcolor[HTML]{FFFFFE}2.27 & \cellcolor[HTML]{FFFEF5}3.53 & \cellcolor[HTML]{FFFEF4}4.18 \\
\textbf{IDR}  & \cellcolor[HTML]{FFF09D}7.09 & \cellcolor[HTML]{FFEF9C}7.48 & \cellcolor[HTML]{FFEF9C}7.64 & \cellcolor[HTML]{FFF0A2}7.33 & \cellcolor[HTML]{FFEF9C}6.92 & \cellcolor[HTML]{FFEF9C}7.10 & \cellcolor[HTML]{FFF09D}7.05 & \cellcolor[HTML]{FFF4B6}6.58 \\
\textbf{QRA}  & \cellcolor[HTML]{FFEF9C}7.11 & \cellcolor[HTML]{FFF2A9}6.83 & \cellcolor[HTML]{FFF0A3}7.37 & \cellcolor[HTML]{FFEF9C}7.52 & \cellcolor[HTML]{FFF1A3}6.58 & \cellcolor[HTML]{FFF2AB}6.37 & \cellcolor[HTML]{FFEF9C}7.07 & \cellcolor[HTML]{FFEF9C}7.54 \\
\textbf{SQRA} & \cellcolor[HTML]{FFF3B2}5.99 & \cellcolor[HTML]{FFF5BE}5.81 & \cellcolor[HTML]{FFF7CD}5.48 & \cellcolor[HTML]{FFFBE2}5.05 & \cellcolor[HTML]{FFF4BA}5.43 & \cellcolor[HTML]{FFF5C0}5.33 & \cellcolor[HTML]{FFF7CB}5.21 & \cellcolor[HTML]{FFFBE3}4.85 \\
\textbf{QRM}  & \cellcolor[HTML]{C1E5CA}1.74 & \cellcolor[HTML]{C2E5CB}2.16 & \cellcolor[HTML]{E4F3E8}2.96 & \cellcolor[HTML]{DCF0E2}3.44 & \cellcolor[HTML]{BEE3C8}1.52 & \cellcolor[HTML]{CEEAD6}1.87 & \cellcolor[HTML]{BAE2C5}2.34 & \cellcolor[HTML]{C8E8D0}2.87 \\
\textbf{SQRM} & \cellcolor[HTML]{63BE7B}1.34 & \cellcolor[HTML]{68C07F}1.62 & \cellcolor[HTML]{87CD9A}1.87 & \cellcolor[HTML]{7FC993}1.90 & \cellcolor[HTML]{65BE7C}1.08 & \cellcolor[HTML]{75C58A}1.23 & \cellcolor[HTML]{67BF7E}1.39 & \cellcolor[HTML]{63BE7B}1.27 \\
\textbf{LQRA} & \cellcolor[HTML]{FFFEF4}2.57 & \cellcolor[HTML]{FFFFFD}2.65 & \cellcolor[HTML]{DDF1E2}2.88 & \cellcolor[HTML]{D1EBD8}3.25 & \cellcolor[HTML]{FFFFFB}2.09 & \cellcolor[HTML]{F7FBF8}2.16 & \cellcolor[HTML]{D5EDDC}2.65 & \cellcolor[HTML]{D7EEDD}3.12 \\
\textbf{iQRA} & \cellcolor[HTML]{FFFFFF}2.01 & \cellcolor[HTML]{E9F5EC}2.39 & \cellcolor[HTML]{D3ECDA}2.76 & \cellcolor[HTML]{D9EFDF}3.39 & \cellcolor[HTML]{CEEAD5}1.60 & \cellcolor[HTML]{D7EEDD}1.93 & \cellcolor[HTML]{DBF0E0}2.71 & \cellcolor[HTML]{E3F3E7}3.30 \\
\textbf{iSQRA}  & \cellcolor[HTML]{6AC081}1.37 & \cellcolor[HTML]{63BE7B}1.59 & \cellcolor[HTML]{63BE7B}1.44 & \cellcolor[HTML]{63BE7B}1.43 & \cellcolor[HTML]{63BE7B}1.07 & \cellcolor[HTML]{63BE7B}1.10 & \cellcolor[HTML]{63BE7B}1.34 & \cellcolor[HTML]{64BE7C}1.30
\end{tabular}
}
\end{table}

Table \ref{tab:ACE} compares the {reliability of prediction intervals at four confidence levels: 98\% (1-99\%), 96\% (2-98\%), 90\% (5-95\%), and 80\% (10-90\%).} Several important observations can be made:

\begin{itemize} 
\itemsep = 0pt
\item {SQRM and iSQRA provide the most reliable coverage among the considered methods across all considered prediction interval levels and both markets, with iSQRA achieving the lowest ACE in most cases.}

\item {IDR, QRA and SQRA suffer strongly from too narrow prediction intervals resulting in higher ACE, which may limit their reliability.}

\item {HS and CP provide strong coverage for extreme quantiles corresponding to the 98\% and 96\% PI, but struggle for 80\% and 90\%.}

\item {The kernel estimation related to smoothed QR consistently improves the reliability of probabilistic forecasts compared to their standard QR equivalents.}

\item {The isotonic regularization improves the calibration of QRA and SQRA across all confidence levels and markets.}

\item {iQRA yields a better tail-calibration than LQRA, as observed in the ACE results for 98\% and 96\% prediction intervals in both markets.}

\end{itemize}

{The same conclusions are also valid for the XGB-based prediction. The results for that model are slightly better in terms of coverage and they are reported in the Appendix in Table \ref{tab:ACE:XGB}.}

\begin{table}[tb!]
\centering
\caption{{Prediction Interval Pinball Score (PIPS) for the ensemble of 25 NARX point forecasts and four prediction intervals, across two markets.}}
\label{tab:PIPS}
\scalebox{1}{
\begin{tabular}{c|c|c|c|c|c|c|c|c}
\multicolumn{1}{l}{}  & \multicolumn{4}{c|}{\textbf{Germany}}  & \multicolumn{4}{c}{\textbf{France}} \\
\multicolumn{1}{l|}{\textbf{}} & \textbf{1-99\%} & \textbf{2-98\%} & \textbf{5-95\%}  & \textbf{10-90\%}  & \textbf{1-99\%} & \textbf{2-98\%} & \textbf{5-95\%}  & \textbf{10-90\%}  \\
\textbf{HS} & \cellcolor[HTML]{FFF6C3}0.93 & \cellcolor[HTML]{FFEF9C}1.45 & \cellcolor[HTML]{FFF09D}2.59 & \cellcolor[HTML]{FFF09E}3.93 & \cellcolor[HTML]{FFFADC}0.95 & \cellcolor[HTML]{FFF5C1}1.51 & \cellcolor[HTML]{FFF0A0}2.73 & \cellcolor[HTML]{FFF1A5}4.15 \\
\textbf{CP} & \cellcolor[HTML]{FFF7CC}0.92 & \cellcolor[HTML]{FFF1A4}1.44 & \cellcolor[HTML]{FFEF9C}2.60 & \cellcolor[HTML]{FFEF9C}3.93 & \cellcolor[HTML]{FFFAE0}0.94 & \cellcolor[HTML]{FFF6C5}1.50 & \cellcolor[HTML]{FFEF9C}2.74 & \cellcolor[HTML]{FFEF9C}4.18 \\
\textbf{IDR}  & \cellcolor[HTML]{FFF1A5}0.97 & \cellcolor[HTML]{FFF6C5}1.40 & \cellcolor[HTML]{FFFDF2}2.39 & \cellcolor[HTML]{FFFFFC}3.65 & \cellcolor[HTML]{FFEF9C}1.09 & \cellcolor[HTML]{FFEF9C}1.56 & \cellcolor[HTML]{FFF8D2}2.64 & \cellcolor[HTML]{FFFDEF}3.96 \\
\textbf{QRA}  & \cellcolor[HTML]{FFEF9C}0.99 & \cellcolor[HTML]{FFF3B2}1.42 & \cellcolor[HTML]{FFF6C5}2.50 & \cellcolor[HTML]{FFF6C5}3.81 & \cellcolor[HTML]{FFF1A4}1.07 & \cellcolor[HTML]{FFF2A9}1.54 & \cellcolor[HTML]{FFF1A9}2.72 & \cellcolor[HTML]{FFF1A8}4.15 \\
\textbf{SQRA} & \cellcolor[HTML]{FFF09E}0.99 & \cellcolor[HTML]{FFF3B3}1.42 & \cellcolor[HTML]{FFF8CE}2.48 & \cellcolor[HTML]{FFF7CC}3.79 & \cellcolor[HTML]{FFF1A7}1.06 & \cellcolor[HTML]{FFF2AF}1.53 & \cellcolor[HTML]{FFF4B9}2.69 & \cellcolor[HTML]{FFF4B7}4.11 \\
\textbf{QRM}  & \cellcolor[HTML]{ABDCB8}0.77 & \cellcolor[HTML]{BAE2C4}1.25 & \cellcolor[HTML]{E3F3E8}2.33 & \cellcolor[HTML]{F6FBF8}3.64 & \cellcolor[HTML]{9FD7AE}0.80 & \cellcolor[HTML]{A9DBB6}1.31 & \cellcolor[HTML]{BBE2C5}2.47 & \cellcolor[HTML]{D6EEDD}3.87 \\
\textbf{SQRM} & \cellcolor[HTML]{A8DAB5}0.76 & \cellcolor[HTML]{B7E1C2}1.24 & \cellcolor[HTML]{E1F2E5}2.33 & \cellcolor[HTML]{F2F9F4}3.63 & \cellcolor[HTML]{97D3A7}0.79 & \cellcolor[HTML]{A1D8AF}1.30 & \cellcolor[HTML]{B7E1C2}2.46 & \cellcolor[HTML]{CFEBD6}3.86 \\
\textbf{LQRA} & \cellcolor[HTML]{71C387}0.71 & \cellcolor[HTML]{67BF7E}1.16 & \cellcolor[HTML]{63BE7B}2.18 & \cellcolor[HTML]{63BE7B}3.46 & \cellcolor[HTML]{8DCF9F}0.78 & \cellcolor[HTML]{76C58B}1.25 & \cellcolor[HTML]{67BF7E}2.36 & \cellcolor[HTML]{66BF7E}3.75 \\
\textbf{iQRA} & \cellcolor[HTML]{6AC181}0.71 & \cellcolor[HTML]{6BC181}1.16 & \cellcolor[HTML]{72C488}2.20 & \cellcolor[HTML]{75C58A}3.49 & \cellcolor[HTML]{70C386}0.76 & \cellcolor[HTML]{6DC284}1.24 & \cellcolor[HTML]{6EC284}2.37 & \cellcolor[HTML]{70C386}3.76 \\
\textbf{iSQRA}  & \cellcolor[HTML]{63BE7B}0.70 & \cellcolor[HTML]{63BE7B}1.15 & \cellcolor[HTML]{6AC181}2.19 & \cellcolor[HTML]{6EC284}3.48 & \cellcolor[HTML]{63BE7B}0.75 & \cellcolor[HTML]{63BE7B}1.23 & \cellcolor[HTML]{63BE7B}2.36 & \cellcolor[HTML]{63BE7B}3.75
\end{tabular}
}

\end{table}

Table~\ref{tab:PIPS} presents the prediction interval pinball scores (PIPS) for {the same four confidence levels}. The table helps identify models that balance sharpness and reliability most effectively.

The {iSQRA and iQRA models deliver the best results for the widest 98\% and 96\% PI in both markets highlighting their strength in generating accurate prediction for extreme quantiles, with slight advantage for smoothed version of the model}. At the {90\% and 80\%} levels, LQRA slightly outperforms {iQRA and iSQRA}, suggesting that it is better suited for sharper intervals where a narrower coverage range is acceptable. {The other methods fall notably behind, with QRM and SQRM being the only ones providing competitive results.}

\subsection{{Economic value of probabilistic forecasts}}
\label{ssec:economic_value}

{Statistical scores such as the CRPS assess the quality of probabilistic forecasts, but they do not directly show whether the forecasts are useful in supporting trading decisions. Therefore, we complement the statistical analysis with a simple trading strategy based on day-ahead arbitrage. {Note that we report this trading evaluation for the German market only. Under the assumed battery efficiency and operating cost parameters, the French market generated too few profitable arbitrage opportunities to allow for a meaningful comparison of forecasting methods.}

Following \citep{mac:lip:uni:26} we consider a battery energy storage system that performs at most one charge--discharge cycle per day. For each day $d$, the trader derives the buying and selling hours by selecting the two hours with the highest difference in the median forecast:
{
\begin{equation}
\hat{S}_{d} = \eta \hat{P}_{d,h_s}^{0.5} - \frac{1}{\eta}\hat{P}_{d,h_b}^{0.5} - C
\end{equation}}
where $h_b$ and $h_s$ are buying and selling hours, respectively, $\eta$ denotes battery efficiency and $C$ is the operating cost. Following \citet{ser:wer:25}, we set {the battery efficiency to $\eta = 0.9$} and the cost of one complete charge--discharge cycle to $C=100$ EUR. This value is intended to approximate the cost of operating a utility-scale lithium-ion battery, including battery degradation and operating expenses. In particular, \citet{ser:wer:25} motivate this level using an approximate battery cell cost of 300 EUR/kWh, scaled to a 1 MWh storage system and divided by an assumed lifetime of 3000 charge--discharge cycles. 

The above strategy can be implemented using point forecasts only, in which case the decision to trade is based solely on the predicted spread, i.e. when $\hat{S}_d>0$. We use this approach as a benchmark. Probabilistic forecasts provide additional information about the uncertainty of the future price, allowing the trader to control risks of the loss.

For the selected buying and selling hours, we quantify the uncertainty by the mean upper and lower partial deviations from the median forecast. More specifically, for the buying hour we average the distances between the median and all upper quantiles, while for the selling hour we average the distances between the median and all lower quantiles. The resulting risk measure $R_d \in\mathbb{R}^+$ is given by the sum of mean upper partial deviations for the buying hour and mean lower partial deviation for the selling hour:
{
\begin{equation}
    R_d = \frac{\eta}{49}\sum_{i=1}^{49}\left(\hat{P}_{d,h_s}^{0.5}-\hat{P}_{d,h_s}^{i/100}\right) + \frac{1}{49\eta}\sum_{i=1}^{49}\left(\hat{P}_{d,h_b}^{1-i/100}-\hat{P}_{d,h_b}^{0.5}\right)
\end{equation}}
A trade is executed if the predicted spread is sufficiently large relative to the associated uncertainty:
\begin{equation}
\hat{S}_{d}>\pi R_d,
\end{equation}
where $\pi$ is the risk threshold. In the empirical study, we consider $\pi \in \{1/3, 2/3, 1\}$. Larger values of $\pi$ correspond to a more risk-averse strategy.

If the condition is satisfied, the realized profit for day $d$ is computed using the observed prices:
{
\begin{equation}
S_d=\eta P_{d,h_s}-\frac{1}{\eta}P_{d,h_b}-C
\end{equation}
}
If the condition is not satisfied, no transaction is made and the daily profit is set to zero. Since different forecasting methods may generate different numbers of trades, we evaluate their economic performance using the average gain per trade rather than the cumulative profit.} {Additionally, we also consider the ability of risk-informed strategies to avoid losses. For this purpose we define loss avoidance $L_A$ as:
\begin{equation}
    L_A = \frac{\sum_{d\in\mathcal{D}} \mathbf{1}_{S_d \geq 0} \mathbf{1}_{\tilde{S}_d < 0}}{\sum_{d\in\mathcal{D}} \mathbf{1}_{\tilde{S}_d < 0}},
\end{equation}
where $S_d$ is the daily profit of the risk-informed strategy and $\tilde{S}_d$ is that of the strategy based solely on point forecasts. The measure gives the proportion of loss-making days for the point-forecast strategy on which the risk-informed strategy breaks even or earns a positive profit.
}

\begin{figure}
\caption{{Gain per trade (EUR) and loss avoidance (\%) from strategies based on postprocessed ensemble of 25 NARX forecasts, on the German market, aggregated for the entire sample period. Baseline profit per trade equates to 43.1 EUR and is marked by a dashed line. The bars represent strategies with different risk thresholds - $1/3$ for dark blue, $2/3$ for blue, and $1$ for light blue color.}}
\includegraphics[width=0.95\linewidth]{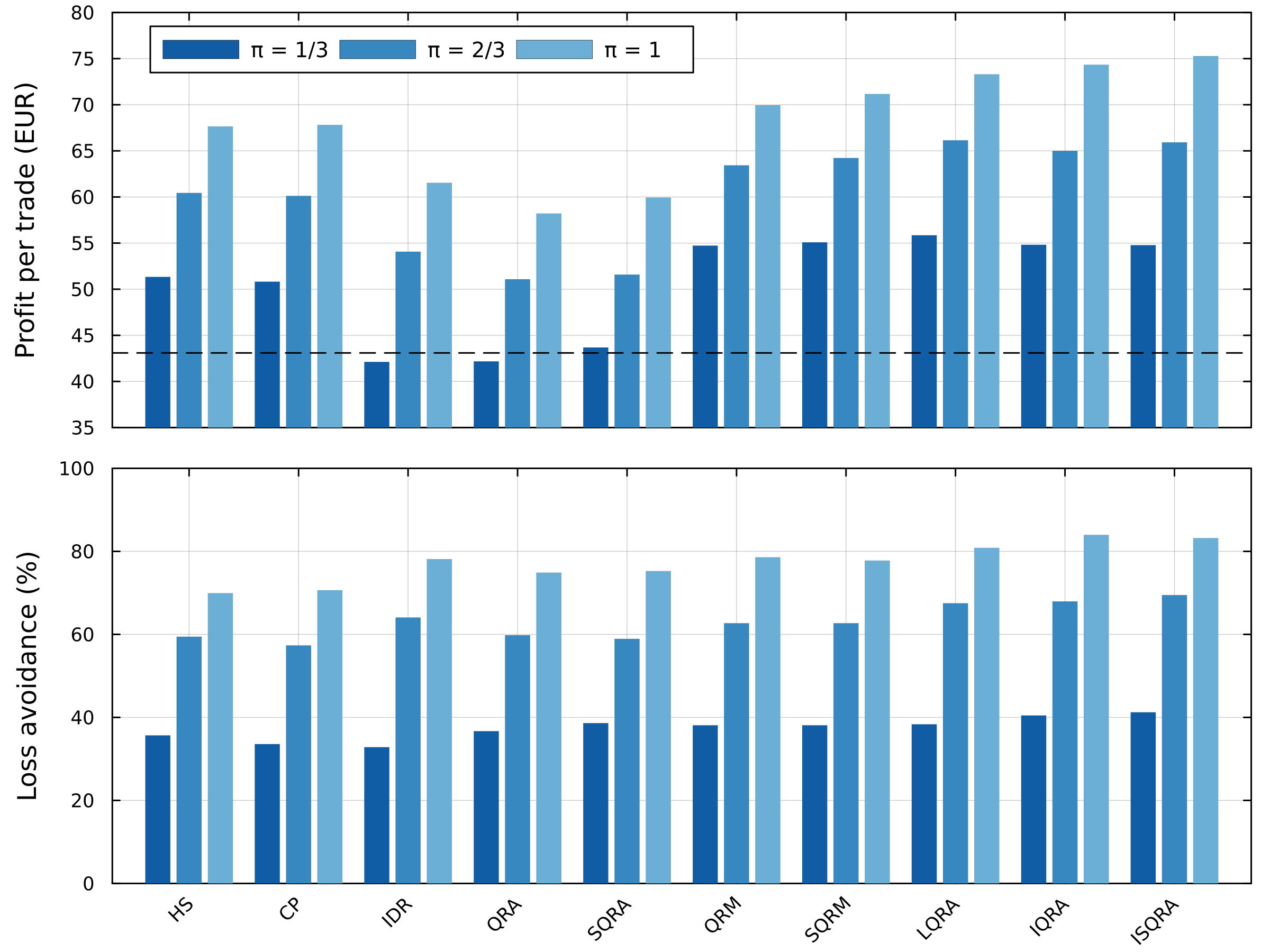}
\label{fig:Trading}
\end{figure}

Figure~\ref{fig:Trading} summarizes the average gain per trade {and loss avoidance} achieved by the considered probabilistic forecasting methods for different values of the risk threshold $\pi$. We report results only for the German market. Under the adopted assumptions on battery efficiency and operating costs, arbitrage opportunities on the French market were very limited, resulting in only a few trades per year. Consequently, the average gains were dominated by random variation and did not allow for a meaningful comparison of forecasting methods.

Several observations can be made from Fig.~\ref{fig:Trading}. First, incorporating probabilistic forecasts into the trading decision consistently improves the average gain per trade compared to the benchmark strategy based solely on point forecasts, indicated by the dashed horizontal line. Second, increasing the risk threshold leads to higher average profits per trade for all methods, reflecting the fact that more conservative strategies execute fewer but better justified trades. {The same conclusion holds for loss avoidance -- the higher the risk threshold the better the ability of the strategy to avoid loss-making trades.}

Finally, the economic evaluation is consistent with the statistical results reported in the previous sections. The methods that achieved the lowest CRPS values, namely LQRA, iQRA, and iSQRA, also generate the highest gains per trade across all considered risk thresholds. In contrast, the simpler benchmark methods, particularly IDR, QRA, and SQRA, yield substantially lower profits. These results demonstrate that improvements in probabilistic forecast quality translate into economically valuable trading decisions. {Note that the ranking does not necessarily coincide with the total profit, since total profit also depends on the number of executed trades reported in \ref{app:sec:trading}.}

\subsection{Regularization performance: Isotonic or Lasso}

{The empirical results show that LQRA, iQRA, and iSQRA form the leading group of quantile-regression-based postprocessing methods, with very similar CRPS values across markets, ensemble sizes, and base forecasting models. Apart from the stronger coverage reliability of iSQRA, the main distinction between lasso-based and isotonic regularization is therefore not a clear difference in aggregate predictive accuracy, but their algorithmic structure. LQRA controls overfitting through an explicit $\ell_1$ penalty and requires selecting a regularization parameter, whereas iQRA and iSQRA impose stochastic-order constraints directly through nonnegative coefficients.

In this section, we therefore focus on the practical advantages of isotonic regularization beyond aggregate accuracy: lower computational cost, improved scalability, and an implicit variable-selection effect without hyperparameter tuning.}

\paragraph{Computational costs}
{
A standard approach to estimating the coefficients of quantile regression is to reformulate the optimization problem in Eq.~\ref{eq:qrmin} as a linear program by introducing $2T$ slack variables~\citep{koe:2005}. In the $\min_{Ax=b, x\geq 0}c^\text{T}x$ form, the total number of variables $n$ is $2(M+T+1)$ and the number of constraints $d$ is $T$. Introducing isotonic regularization decreases the number of variables by $M$, leading to $n= M + 2(T+1)$. {The computational complexity of linear programming algorithms depends on the problem dimensions and, for some algorithms, the number of nonzero entries in the constraint matrix~\citep{lee:sid:2015,bra:2020}. The sizes of corresponding linear programs are compared in Table~\ref{tab:complexity}. By reducing both the number of variables and the number of nonzero entries, isotonic regularization can therefore reduce computation time. In contrast, regularizing with lasso increases the overall computation time, as multiple optimization problems have to be solved for different values of $\lambda$ in order to select the optimal regularization strength. In practice, we solve linear programs with the open-source HiGHS software~\citep{hu:ha:2018} and allow for automatic algorithm selection. The computation time for each of quantile regression methods we considered in this paper are presented in Table~\ref{tab:time}. Note that the time required by LQRA depends on the size of the considered grid of $\lambda$ values. The practical time evaluations indicate that iQRA provides increasing efficiency gain the larger the forecast ensemble we postprocess, i.e. the higher the number of regressors $M$.}

Isotonic regularization also reduces the computational time for the family of smoothed quantile regressions, which we compute with the Sequential Least Squares Quadratic Programming (SLSQP) method~\citep{kra:94} from the NLopt library~\citep{NLopt}. Limiting the search space to non-negative parameter values and initializing the optimization with the isotonic solution can provide significant speedup for postprocessing ensembles {with smoothed quantile regression}, from around 45\% to 85\% depending on the ensemble size.
}

\begin{table}[]
    \centering
    
    \begin{tabular}{l|c|c}
        \hline
        Quantity & QRA & iQRA \\
        \hline
        Nonnegative variables & $2M+2T+2$ & $M+2T+2$ \\
        Equality constraints & $T$ & $T$ \\
        Nonzero entries in $A$ & $T(2M+4)$ & $T(M+4)$ \\
        \hline
    \end{tabular}
    \caption{Sizes of the QRA and iQRA linear programming formulations
    in the standard form $\min_{Ax=b,\,x\geq 0} c^\top x$,
    where $M$ is the number of regressors and $T$ is the number
    of observations. Both formulations include an unrestricted
    intercept, while iQRA constrains the slope coefficients to be
    nonnegative. The nonzero-entry counts assume that all entries
    of the regressor matrix are nonzero.}
    \label{tab:complexity}
\end{table}

\begin{table}[b!]
\centering
\caption{{Time (s) required for each quantile regression method to generate quantile forecasts for a single day (99 quantiles for each of the 24 hours) in Julia with a single execution thread on Apple M2 Pro, not including the compilation time. The values include average and standard deviation over 10 runs.} {$\textbf{LQRA}_{20}$ corresponds to the time for estimating the model over the full regularization path consisting of 20 $\lambda$ values, while $\textbf{LQRA}_{1}$ indicates the time per single $\lambda$ value.}}
\label{tab:time}
\begin{tabular}{c|c|c|c}
\textbf{QRM}  & \multicolumn{3}{c}{3.03 $\pm$ 0.02} \\
\textbf{SQRM} & \multicolumn{3}{c}{3.76 $\pm$ 0.04} \\
\hline
\multicolumn{1}{l|}{Ensemble size} & 10  & 25 & 50  \\
\hline
\textbf{QRA}  & 4.56 $\pm$ 0.23  & 7.56 $\pm$ 0.15 & 11.73 $\pm$ 0.13 \\
\textbf{SQRA} & 14.78 $\pm$ 0.37 & 96.71 $\pm $2.40 & 473.93 $\pm$ 16.02  \\
{$\textbf{LQRA}_{20}$} & 117.98 $\pm$ 2.13  & 220.40 $\pm$ 8.45 & 456.14 $\pm$ 22.96 \\
{$\textbf{LQRA}_{1}$} & 5.90 $\pm$ 0.11  & 11.02 $\pm$ 0.42 & 22.81$\pm$ 1.15 \\
\textbf{iQRA} & 3.83 $\pm$ 0.13  & 4.20 $\pm$ 0.08 & 5.32 $\pm$ 0.23  \\
\textbf{iSQRA}  & 8.05 $\pm$ 0.38  & 22.96 $\pm$ 0.98  & 69.24 $\pm$ 3.59  
\end{tabular}
\\
\end{table}

\paragraph{Variable selection} 

\begin{figure}[b!]
 \includegraphics[width = 0.85\linewidth]{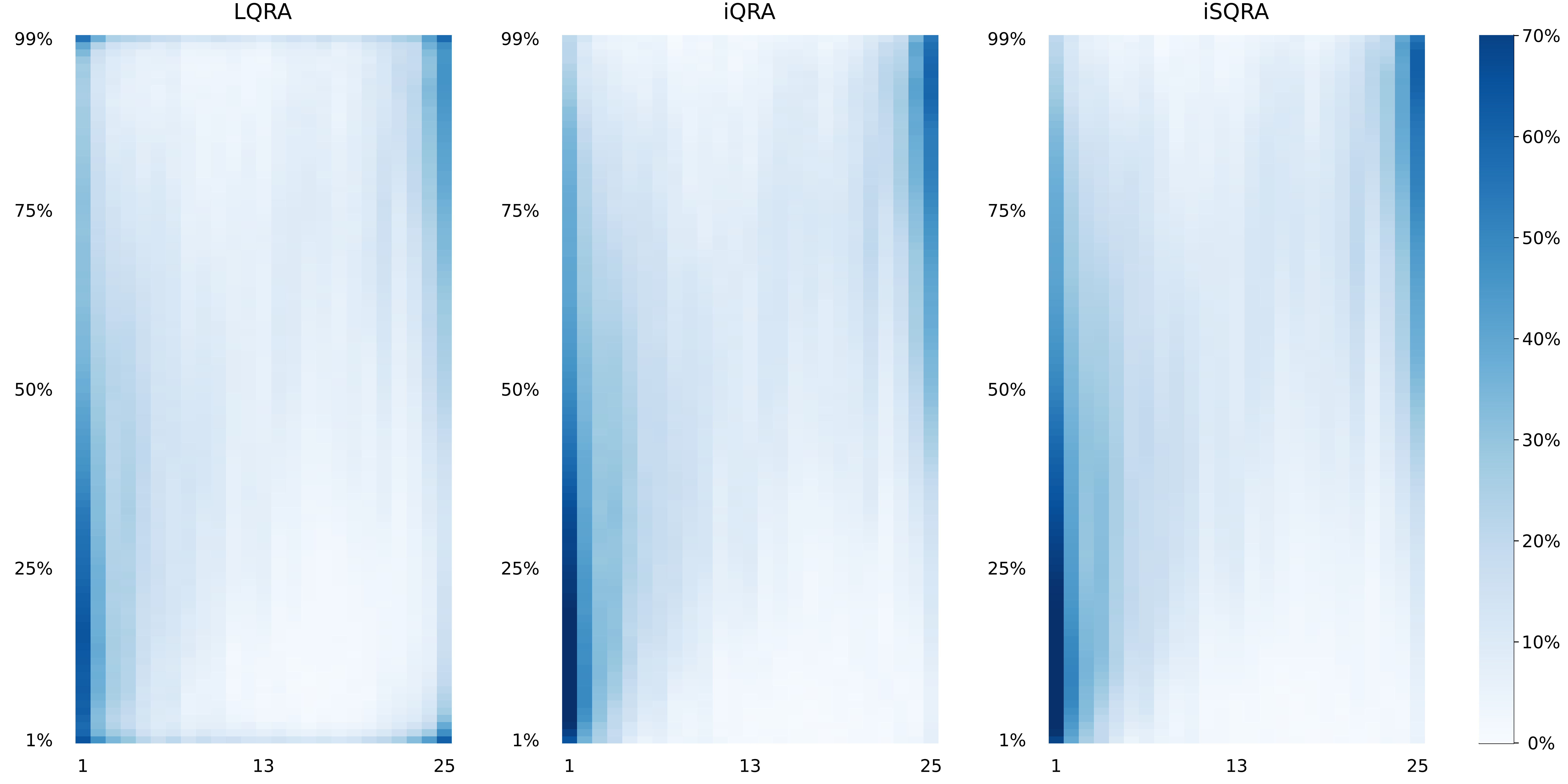}
 \caption{The plots show how often (in percentages) given predictions are included in the final model (corresponding $\beta_i$ has non-zero value). We report the percentages separately for each forecasted quantile (vertical axis) and for each point prediction in the ensemble (horizontal axis). Recall that the 1-st prediction corresponds to the ensemble minimum, 13-th to the median, and 25-th to the maximum. The aggregated percentage of selected variables is 11.4\% for LQRA, 13.7\% for iQRA and 14.1\% for iSQRA. {The plots correspond to the ensemble of 25 NARX forecast in the German market.}}
 \label{fig:Frequency}
\end{figure}

In Fig.~\ref{fig:Frequency} we show how often a given {ensemble member} was selected in the final model (corresponding $\beta_i$ value was different from 0). The results are presented for the iQRA, {iSQRA} and LQRA models. The darker the color in Fig.~\ref{fig:Frequency}, the more often the given {ensemble member (on the X-axis)} was selected as important to forecast a quantile at the given probability level {(Y-axis)}. It can be seen that by far the darkest places are on the left and right side of both plots, indicating that the extreme predictions are selected much more often than the middle ones. Another observation is that the smallest prediction is more often selected to predict lower quantiles, whereas the highest predictions are more often selected to obtain the forecast for upper part of the distribution. Interestingly LQRA, unlike iQRA and iSQRA, also selects the smallest forecast of the ensemble to predict the upper tail, and the highest forecast to predict the lower tail quantiles. {Despite using information from both tails, LQRA selects fewer predictors in total, when compared to iQRA and iSQRA.} It follows from this analysis that regularization through isotonicity also performs variable selection, effectively reducing the number of regressors. At the same time, it does not require any hyperparameter to tune the intensity of regularization. Since the assumption of stochastic order typically requires prior expert knowledge about the underlying processes~\citep{tak:etal:2006}, isotonic linear regression quantiles can be potentially used as an automatic method for selecting significant isotonic regressors for other models. We leave this problem for further research.

\section{Conclusion}
\label{sec:Conclusions}
This paper introduces Isotonic Quantile Regression Averaging (iQRA) {and its smoothed counterpart, iSQRA,} as robust methods for probabilistic forecasting of electricity prices, particularly in the context of ensemble-based prediction frameworks. By imposing isotonic constraints within the quantile regression setting, the proposed methods enhance the estimation of predictive distributions in a computationally efficient and interpretable manner.

The empirical analysis {on the German and French day-ahead electricity markets} demonstrates that {iQRA and iSQRA} provide reliable and sharp prediction intervals across a range of confidence levels. {In particular, iSQRA achieves the best reliability in terms of coverage, while both methods attain CRPS values comparable to the regularized LQRA benchmark. The results are consistent across different ensemble sizes, and additional experiments with XGB ensembles, reported in the Appendix, lead to the same conclusions, indicating that the proposed methods are robust to the choice of the underlying point forecasting model.} Furthermore, {isotonic regularization eliminates the need for hyperparameter tuning and substantially reduces the computational cost compared with LQRA, making iQRA and iSQRA efficient and scalable solutions for practical deployment.} {Finally, a simple battery trading study demonstrates that the improved probabilistic forecasts translate into competitive, and in some cases slightly higher, economic value than the regularized benchmark {in terms of gain per trade and loss avoidance.}}

Beyond their empirical performance, {iQRA and iSQRA} address a key challenge in the deployment of machine learning models such as NARX {and XGB}: the need to quantify uncertainty around point forecasts. AI-based methods are increasingly used in critical decision-making contexts, yet their deterministic outputs often lack insight into predictive reliability. {The proposed methods bridge this gap by producing well-calibrated predictive distributions without sacrificing computational efficiency, allowing users to better assess risk and support operational decision-making.}

{Several limitations also suggest directions for future research. First, the present study considers rolling-window batch estimation of quantile-regression-based postprocessing models. Considering online or sequential training and regularization schemes could further reduce the computational burden of daily re-estimation and provide a useful extension of the proposed framework. Second, although the robustness checks include forecast ensembles obtained from both feedforward neural networks and gradient boosted trees, each ensemble is generated within a single model family. Applying isotonic regularization to more heterogeneous multi-model ensembles would provide an important test of its usefulness when the set of base forecasts contains more diverse information. Future research directions may also include applying the proposed isotonic quantile regression framework to ensemble postprocessing in other sectors and exploring its role in automatically identifying isotonic regressors within broader modeling frameworks.}

\section*{Declaration of competing interests}
The authors declare no competing interests.

\section*{Acknowledgements}
 The study was partially supported by the National Science Center (NCN, Poland) through grant no.\ 2018/30/A/HS4/00444 (to AL) and grant no.\ 2023/49/N/HS4/02741 (to BU).
 BU is supported by the Foundation for Polish Science (FNP).

\section*{Code and data availability}
The data used for training the neural networks cannot be shared due to terms and conditions of access, but is publicly available at (\url{transparency.entsoe.eu}) and \url{investing.com}. The point forecasts used for postprocessing and Julia source code used in this study are available at \url{github.com/lipiecki/iqra}. All postprocessing methods were computed with the Julia package \textit{PostForecasts.jl}.

\FloatBarrier
\bibliographystyle{apalike} 
\bibliography{ref}

\appendix
\section{{Point forecast accuracy}}
\label{app:point_forecasts}

\begin{table}[H]
\caption{{Mean Absolute Error (MAE) for point forecasts obtained as an averaged prediction of NARX or XGB ensembles, for ensemble sizes of 10, 25 and 50 models. MAE is reported separately for each of the six calendar years in the out-of-sample period, on German and French markets.}}
\centering
\scalebox{0.95}{\begin{tabular}{c|c|c|c|c|c|c|c|c|c|c|c|c}
\multicolumn{1}{l}{}  & \multicolumn{6}{c|}{\textbf{Germany}} & \multicolumn{6}{c}{\textbf{France}}  \\
\multicolumn{1}{l|}{\textbf{}} & \textbf{2020}  & \textbf{2021}  & \textbf{2022} & \textbf{2023}  & \textbf{2024}  & \textbf{2025}  & \textbf{2020}  & \textbf{2021}  & \textbf{2022} & \textbf{2023}  & \textbf{2024}  & \textbf{2025}  \\
\textbf{NARX 10}     & 4.07          & 11.89         & 29.24         & 14.85         & 13.48         & 14.24         & 3.89          & 13.74         & 32.86         & 15.22         & 14.32         & 14.25         \\
\textbf{NARX 25}     & 4.02          & 11.76         & 29.00         & 14.72         & 13.37         & 14.13         & 3.85          & 13.63         & 32.46         & 15.07         & 14.24         & 14.15         \\
\textbf{NARX 50}     & 4.00          & 11.72         & 28.89         & 14.64         & 13.33         & 14.10         & 3.83          & 13.58         & 32.46         & 15.03         & 14.22         & 14.13         \\
\textbf{XGB 10}      & 4.17          & 14.50         & 32.41         & 13.91         & 12.74         & 12.72         & 4.07          & 15.29         & 35.63         & 14.81         & 14.09         & 13.90         \\
\textbf{XGB 25}      & 4.15          & 14.47         & 32.31         & 13.81         & 12.71         & 12.70         & 4.02          & 15.24         & 35.51         & 14.71         & 14.05         & 13.83         \\
\textbf{XGB 50}      & 4.15          & 14.45         & 32.25         & 13.80         & 12.70         & 12.67         & 4.02          & 15.24         & 35.51         & 14.72         & 14.04         & 13.80        
\end{tabular}
}
\end{table}

\section{{Trading strategy}}
\label{app:sec:trading}
\begin{table}[H]
\caption{{Number of trades and profit per trade for the economic evaluation of the postprocessed ensemble of 25 NARX models in the German day-ahead market. Profit per trade corresponds to the values plotted in Fig.~\ref{fig:Trading}}.}
\centering
\begin{tabular}{c|c|c|c|c|c|c}
\multicolumn{1}{c}{} & \multicolumn{3}{c|}{\textbf{Number of trades}} & \multicolumn{3}{c}{\textbf{Profit per trade}} \\
$\pi$ & $1/3$ & $2/3$ & $1$ & $1/3$ & $2/3$ & $1$ \\
\hline
\textbf{HS}                   & 535           & 433           & 357           & 51.3          & 60.4          & 67.6          \\
\textbf{CP}                   & 539           & 429           & 355           & 50.8          & 60.1          & 67.8          \\
\textbf{IDR}                  & 586           & 444           & 337           & 42.1          & 54.1          & 61.5          \\
\textbf{QRA}                  & 550           & 449           & 348           & 42.2          & 51.1          & 58.2          \\
\textbf{SQRA}                 & 545           & 449           & 343           & 43.7          & 51.6          & 59.9          \\
\textbf{QRM}                  & 485           & 389           & 297           & 54.7          & 63.4          & 69.9          \\
\textbf{SQRM}                 & 483           & 382           & 288           & 55.1          & 64.2          & 71.2          \\
\textbf{LQRA}                 & 477           & 368           & 285           & 55.8          & 66.1          & 73.3          \\
\textbf{iQRA}                 & 492           & 382           & 299           & 54.8          & 65.0          & 74.3          \\
\textbf{iSQRA}                & 488           & 376           & 291           & 54.8          & 65.9          & 75.3       
\end{tabular}
\end{table}

\section{{Postprocessing XGB forecasts}}
\label{app:sec:xgb}
{To examine whether the benefits of isotonic regularization for (S)QRA extend beyond neural networks we consider an ensemble of gradient-boosted decision trees~\citep{has:tib:fri:09}. This widely adopted machine learning method demonstrated top performance in the M5 forecasting competition~\citep{jan:etal:2022} and represents a different approach to function approximation than neural networks. We use the eXtreme Gradient Boosting (XGB) implementation from the Python XGBoost package~\citep{che:gue:16}, with the mean squared error (MSE) as the loss function. The hyperparameters of the XGB model are selected independently for each hour and ensemble member, using Bayesian sequential optimization \citep{optuna_2019}. The hyperparameter tuning is conducted before each calendar year, on the same training window that is used for predicting the first day of the year. The evaluation is based on a random 20\% of the training window, held out for validation. Each optimization study comprises 1000 consecutive trials. Hyperparameters included in the optimization and the respective ranges of values are provided in Table~\ref{tab:hyperparams}. We use the same set of predictors and training window size as for the NARX model.}

\begin{table}[h!]
    \centering
    \begin{tabular}{c|c}
         Parameter  & Range of values\\
         \hline
         Maximum depth & $\{2,..., 10\}$ \\
         Learning rate & $[10^{-4}, 1]^\dagger$ \\
         Subsampling rate & $[0.5, 1]$ \\
         Minimum child weight & $\{0, ..., 10\}$ \\
         Minimum loss reduction ($\gamma$) & $[0, 5]$ \\
         L1 regularization strength ($\lambda$) & $[10^{-3}, 10]^\dagger$\\
         L2 regularization strength ($\alpha$) & $[10^{-3}, 10]^\dagger$
    \end{tabular}
    \caption{Hyperparameters and the considered value ranges for the XGB model. Logarithmic scale was used for the ranges indicated with $\dagger$.}
    \label{tab:hyperparams}
\end{table}

\begin{table}[h!]
\centering
\caption{Continuous ranked probability score (CRPS) for a postprocessed ensemble of 25 XGB point forecasts and {six} years of out-of-sample period on German and French markets.}
\label{tab:CRPS:XGB}
\scalebox{1}{
\begin{tabular}{c|c|c|c|c|c|c|c|c|c|c|c|c}
\multicolumn{1}{l}{}  & \multicolumn{6}{c|}{\textbf{Germany}} & \multicolumn{6}{c}{\textbf{France}}  \\
\multicolumn{1}{l|}{\textbf{}} & \textbf{2020}  & \textbf{2021}  & \textbf{2022} & \textbf{2023}  & \textbf{2024}  & \textbf{2025}  & \textbf{2020}  & \textbf{2021}  & \textbf{2022} & \textbf{2023}  & \textbf{2024}  & \textbf{2025}  \\
\textbf{HS} & \cellcolor[HTML]{F2F9F4}1.56 & \cellcolor[HTML]{FFF4BA}5.69 & \cellcolor[HTML]{FFFBE2}12.26 & \cellcolor[HTML]{FFEF9C}5.57 & \cellcolor[HTML]{FFFCE8}4.92 & \cellcolor[HTML]{FFF8D1}4.80 & \cellcolor[HTML]{FFFFFE}1.50 & \cellcolor[HTML]{FFF5BC}6.00 & \cellcolor[HTML]{FFFCED}13.63 & \cellcolor[HTML]{FFEF9C}5.95 & \cellcolor[HTML]{FAFDFB}5.06 & \cellcolor[HTML]{FFFFFB}5.03 \\
\textbf{CP} & \cellcolor[HTML]{E6F4E9}1.56 & \cellcolor[HTML]{FFEF9C}5.87 & \cellcolor[HTML]{FFFBE5}12.24 & \cellcolor[HTML]{FFF2A9}5.56 & \cellcolor[HTML]{FFFFFC}4.88 & \cellcolor[HTML]{FFF6C5}4.84 & \cellcolor[HTML]{FFF8D3}1.52 & \cellcolor[HTML]{FFEF9C}6.16 & \cellcolor[HTML]{FFFCE8}13.67 & \cellcolor[HTML]{FFF1A7}5.92 & \cellcolor[HTML]{FFFBE2}5.13 & \cellcolor[HTML]{FFF9D9}5.12 \\
\textbf{IDR}  & \cellcolor[HTML]{FFF2AA}1.64 & \cellcolor[HTML]{FFF8D4}5.54 & \cellcolor[HTML]{FFEF9C}12.68 & \cellcolor[HTML]{63BE7B}5.20 & \cellcolor[HTML]{FFF4B7}5.01 & \cellcolor[HTML]{9FD7AD}4.61 & \cellcolor[HTML]{FFFADE}1.51 & \cellcolor[HTML]{FFF5C1}5.98 & \cellcolor[HTML]{FFF4B8}14.05 & \cellcolor[HTML]{DDF1E2}5.65 & \cellcolor[HTML]{FFFADB}5.15 & \cellcolor[HTML]{78C78D}4.97 \\
\textbf{QRA}  & \cellcolor[HTML]{FFEF9C}1.65 & \cellcolor[HTML]{FFFCE8}5.42 & \cellcolor[HTML]{FFF09E}12.67 & \cellcolor[HTML]{FFFBE3}5.48 & \cellcolor[HTML]{FFEF9C}5.06 & \cellcolor[HTML]{FFEF9C}4.95 & \cellcolor[HTML]{FFEF9C}1.54 & \cellcolor[HTML]{FFF9D8}5.87 & \cellcolor[HTML]{FFEF9C}14.27 & \cellcolor[HTML]{FFF3B3}5.89 & \cellcolor[HTML]{FFEF9C}5.31 & \cellcolor[HTML]{FFEF9C}5.28 \\
\textbf{SQRA} & \cellcolor[HTML]{FFF0A3}1.64 & \cellcolor[HTML]{FFFCE7}5.42 & \cellcolor[HTML]{FFF1A5}12.63 & \cellcolor[HTML]{FFFBE3}5.48 & \cellcolor[HTML]{FFF2AA}5.03 & \cellcolor[HTML]{FFF0A1}4.93 & \cellcolor[HTML]{FFF2AA}1.53 & \cellcolor[HTML]{FFF9D5}5.88 & \cellcolor[HTML]{FFF1A3}14.22 & \cellcolor[HTML]{FFF3B3}5.89 & \cellcolor[HTML]{FFF2AC}5.27 & \cellcolor[HTML]{FFF1A6}5.25 \\
\textbf{QRM}  & \cellcolor[HTML]{FFFFFB}1.57 & \cellcolor[HTML]{B9E2C4}5.14 & \cellcolor[HTML]{BBE2C5}11.92 & \cellcolor[HTML]{F6FBF7}5.42 & \cellcolor[HTML]{EFF8F2}4.86 & \cellcolor[HTML]{FBFDFC}4.69 & \cellcolor[HTML]{FDFEFE}1.50 & \cellcolor[HTML]{A1D8B0}5.48 & \cellcolor[HTML]{B7E1C2}13.33 & \cellcolor[HTML]{F7FBF8}5.67 & \cellcolor[HTML]{DCF0E1}5.05 & \cellcolor[HTML]{FBFDFB}5.02 \\
\textbf{SQRM} & \cellcolor[HTML]{FFFFFD}1.57 & \cellcolor[HTML]{B1DEBD}5.12 & \cellcolor[HTML]{B0DEBC}11.90 & \cellcolor[HTML]{FFFEF6}5.45 & \cellcolor[HTML]{EDF7F0}4.86 & \cellcolor[HTML]{FFFFFF}4.70 & \cellcolor[HTML]{FDFEFE}1.50 & \cellcolor[HTML]{9BD5AA}5.47 & \cellcolor[HTML]{A7DAB4}13.30 & \cellcolor[HTML]{FFFFFD}5.69 & \cellcolor[HTML]{C2E5CB}5.05 & \cellcolor[HTML]{FFFFFF}5.02 \\
\textbf{LQRA} & \cellcolor[HTML]{B4DFBF}1.55 & \cellcolor[HTML]{68C080}4.98 & \cellcolor[HTML]{63BE7B}11.72 & \cellcolor[HTML]{68C07F}5.21 & \cellcolor[HTML]{A7DAB4}4.83 & \cellcolor[HTML]{63BE7B}4.56 & \cellcolor[HTML]{73C488}1.44 & \cellcolor[HTML]{66BF7D}5.36 & \cellcolor[HTML]{64BE7C}13.16 & \cellcolor[HTML]{63BE7B}5.53 & \cellcolor[HTML]{FFFFFF}5.06 & \cellcolor[HTML]{63BE7B}4.96 \\
\textbf{iQRA} & \cellcolor[HTML]{75C58A}1.54 & \cellcolor[HTML]{68C07F}4.98 & \cellcolor[HTML]{7AC78E}11.77 & \cellcolor[HTML]{9DD6AC}5.29 & \cellcolor[HTML]{6FC385}4.80 & \cellcolor[HTML]{8FD0A0}4.60 & \cellcolor[HTML]{65BE7C}1.43 & \cellcolor[HTML]{69C080}5.37 & \cellcolor[HTML]{76C68B}13.20 & \cellcolor[HTML]{8BCF9D}5.57 & \cellcolor[HTML]{85CC98}5.03 & \cellcolor[HTML]{73C589}4.96 \\
\textbf{iSQRA}  & \cellcolor[HTML]{63BE7B}1.54 & \cellcolor[HTML]{63BE7B}4.97 & \cellcolor[HTML]{6DC283}11.74 & \cellcolor[HTML]{ABDCB8}5.31 & \cellcolor[HTML]{63BE7B}4.79 & \cellcolor[HTML]{91D1A2}4.60 & \cellcolor[HTML]{63BE7B}1.43 & \cellcolor[HTML]{63BE7B}5.35 & \cellcolor[HTML]{63BE7B}13.15 & \cellcolor[HTML]{9AD5AA}5.58 & \cellcolor[HTML]{63BE7B}5.02 & \cellcolor[HTML]{78C78D}4.97
\end{tabular}
}
\end{table}

\begin{figure}[h!]
  \centering
  \begin{minipage}{0.49\textwidth}
  \centering  
  \textbf{\quad Germany}
  
  \includegraphics[width=\linewidth]{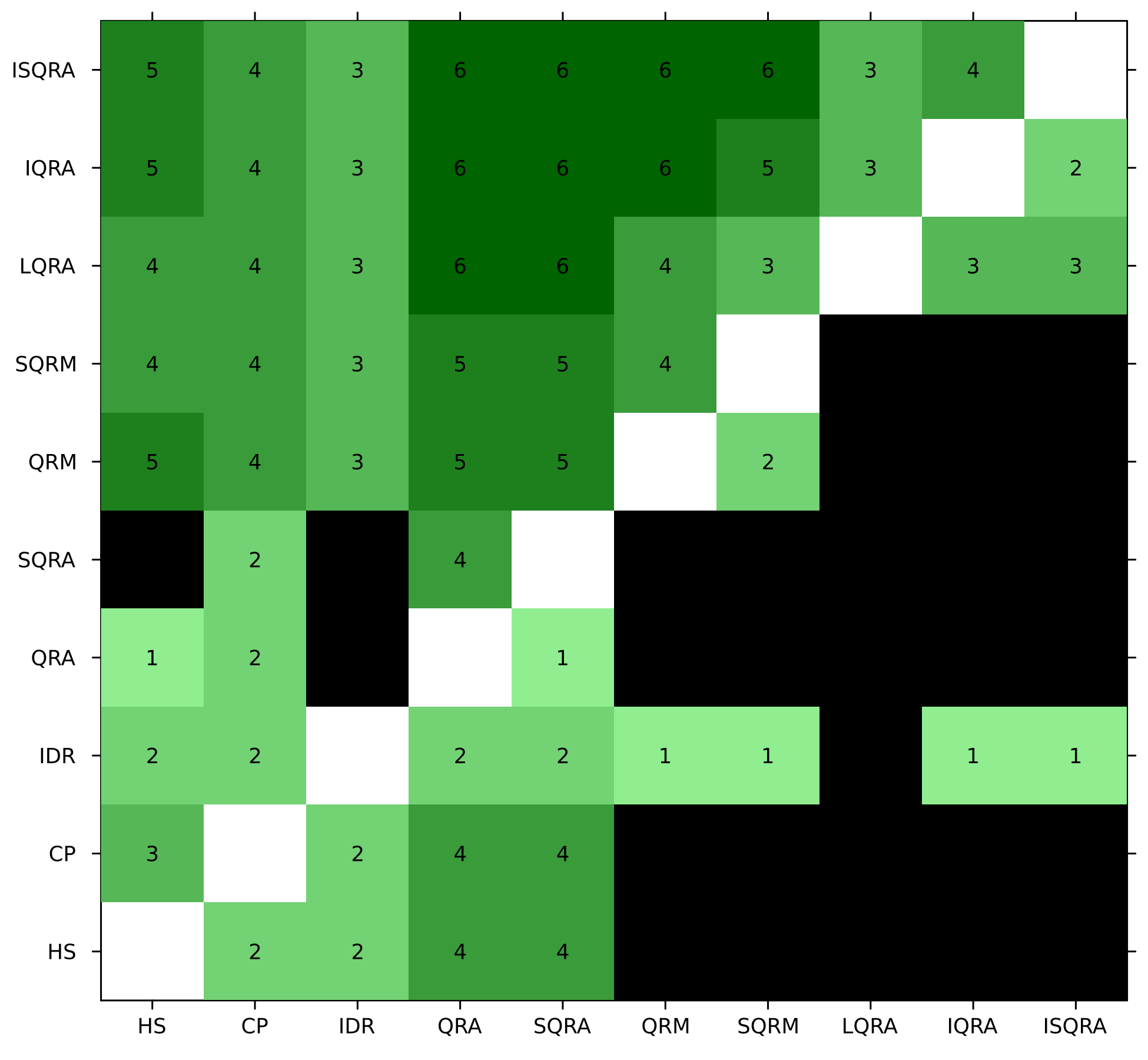}
  \end{minipage}
  \begin{minipage}{0.49\textwidth}
  \centering  
  \textbf{\quad France}
  
  \includegraphics[width=\linewidth]{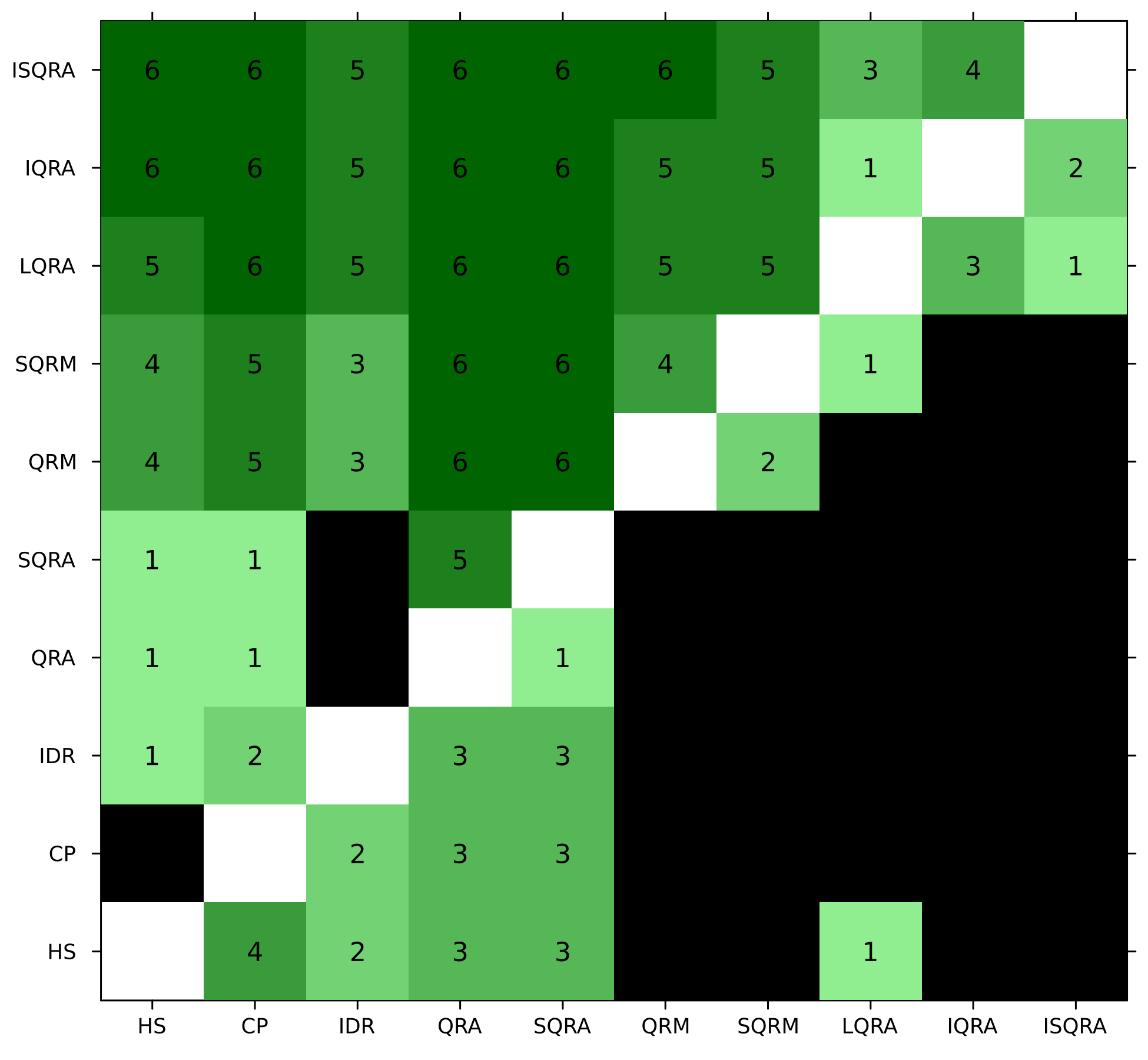}
  \end{minipage}
  \caption{Number of years in which the model on the vertical axis outperformed the model on the horizontal axis according to the Giacomini-White test at the 1\% significance level. Results obtained for an ensemble of 25 XGB forecasts.}
  \label{fig:XGB:tests}
\end{figure}

\begin{table}[h!]
\centering
\caption{Continuous ranked probability score (CRPS) for different sizes of the ensemble of XGB point forecasts, on German and French markets, aggregated for the entire sample period.}
\label{tab:CRPS:XGB:ens}
\scalebox{1}{
\begin{tabular}{c|c|c|c|c|c|c}
\multicolumn{1}{l}{}  & \multicolumn{3}{c|}{\textbf{Germany}} & \multicolumn{3}{c}{\textbf{France}}  \\
\multicolumn{1}{l|}{\textbf{}} & \textbf{10}  & \textbf{25}  & \textbf{50}  & \textbf{10}  & \textbf{25}  & \textbf{50}  \\
\textbf{HS} & \cellcolor[HTML]{FFFBE4}5.818 & \cellcolor[HTML]{FFFBE6}5.800 & \cellcolor[HTML]{FFFCE8}5.792 & \cellcolor[HTML]{FFFCEB}6.220 & \cellcolor[HTML]{FFFDEE}6.194 & \cellcolor[HTML]{FFFDEE}6.192 \\
\textbf{CP} & \cellcolor[HTML]{FFFBE1}5.840 & \cellcolor[HTML]{FFFBE3}5.822 & \cellcolor[HTML]{FFFBE4}5.814 & \cellcolor[HTML]{FFFBE3}6.277 & \cellcolor[HTML]{FFFCE7}6.250 & \cellcolor[HTML]{FFFCE7}6.249 \\
\textbf{IDR}  & \cellcolor[HTML]{FFFCEA}5.775 & \cellcolor[HTML]{FFFCEA}5.777 & \cellcolor[HTML]{FFFCE9}5.784 & \cellcolor[HTML]{FFFCEB}6.219 & \cellcolor[HTML]{FFFCEB}6.215 & \cellcolor[HTML]{FFFCEA}6.226 \\
\textbf{QRA}  & \cellcolor[HTML]{FFFFFD}5.644 & \cellcolor[HTML]{FFFADD}5.867 & \cellcolor[HTML]{FFF09D}6.313 & \cellcolor[HTML]{FFFFFA}6.098 & \cellcolor[HTML]{FFF9D9}6.356 & \cellcolor[HTML]{FFEF9C}6.833 \\
\textbf{SQRA} & \cellcolor[HTML]{FFFFFF}5.631 & \cellcolor[HTML]{FFFADF}5.854 & \cellcolor[HTML]{FFEF9C}6.318 & \cellcolor[HTML]{FFFFFD}6.080 & \cellcolor[HTML]{FFFADC}6.338 & \cellcolor[HTML]{FFF09D}6.829 \\
\textbf{QRM}  & \cellcolor[HTML]{F8FCF9}5.617 & \cellcolor[HTML]{E7F5EB}5.599 & \cellcolor[HTML]{E0F2E4}5.591 & \cellcolor[HTML]{E7F5EB}6.035 & \cellcolor[HTML]{CBE9D3}6.007 & \cellcolor[HTML]{C8E8D0}6.004 \\
\textbf{SQRM} & \cellcolor[HTML]{F6FBF7}5.615 & \cellcolor[HTML]{E5F4E9}5.597 & \cellcolor[HTML]{DEF1E3}5.589 & \cellcolor[HTML]{E1F2E6}6.029 & \cellcolor[HTML]{C5E6CE}6.001 & \cellcolor[HTML]{C2E5CB}5.998 \\
\textbf{LQRA} & \cellcolor[HTML]{94D2A4}5.510 & \cellcolor[HTML]{71C386}5.472 & \cellcolor[HTML]{63BE7B}5.457 & \cellcolor[HTML]{9BD5AA}5.960 & \cellcolor[HTML]{6DC283}5.915 & \cellcolor[HTML]{65BE7C}5.907 \\
\textbf{iQRA} & \cellcolor[HTML]{A6DAB3}5.529 & \cellcolor[HTML]{85CC98}5.494 & \cellcolor[HTML]{77C68C}5.479 & \cellcolor[HTML]{A0D7AE}5.965 & \cellcolor[HTML]{76C68B}5.924 & \cellcolor[HTML]{6BC181}5.913 \\
\textbf{iSQRA}  & \cellcolor[HTML]{A2D8B0}5.525 & \cellcolor[HTML]{81CA95}5.490 & \cellcolor[HTML]{72C488}5.474 & \cellcolor[HTML]{99D4A8}5.958 & \cellcolor[HTML]{6EC284}5.916 & \cellcolor[HTML]{63BE7B}5.905
\end{tabular}
}
\end{table}

\begin{table}[tb!]
\centering

\caption{{The results of the Model Confidence Set based on the daily CRPS values and the $T_{\text{max}}$ statistic~\citep{han:etal:11} for the ensemble of XGB point forecasts. The columns correspond to two markets and three different ensemble sizes. The test was run with 10,000 bootstrap replications and two significance levels: $\alpha = 1\%$ and $\alpha = 5\%$. One checkmark indicates that the model was included in the 99\% confidence set, two checkmarks denote models included in the 95\% confidence set. Models omitted from the table were not retained in either confidence set.}}
\label{tab:MCS:XGB}
\scalebox{1}{
\begin{tabular}{c|c|c|c|c|c|c}
\multicolumn{1}{l}{}  & \multicolumn{3}{c|}{\textbf{Germany}}  & \multicolumn{3}{c}{\textbf{France}} \\
\multicolumn{1}{l|}{\textbf{}} & \textbf{10} & \textbf{25} & \textbf{50}  & \textbf{10}  & \textbf{25} & \textbf{50} \\
\textbf{LQRA} & \checkmark \checkmark  & \checkmark \checkmark  & \checkmark \checkmark & \checkmark \checkmark & \checkmark \checkmark & \checkmark \checkmark \\
\textbf{iQRA}  &  &  & & \checkmark \checkmark & \checkmark & \checkmark \checkmark \\
\textbf{iSQRA} & \checkmark & \checkmark\checkmark & \checkmark \checkmark & \checkmark \checkmark & \checkmark \checkmark & \checkmark \checkmark \\
\end{tabular}
}
\end{table}

\begin{table}[h!]
\caption{Average Coverage Error (\%) for the ensemble of 25 XGB point forecasts and four prediction intervals, across two markets.}
\label{tab:ACE:XGB}
\scalebox{1}{
\begin{tabular}{c|c|c|c|c|c|c|c|c}
\multicolumn{1}{l}{}  & \multicolumn{4}{c|}{\textbf{Germany}}  & \multicolumn{4}{c}{\textbf{France}} \\
\multicolumn{1}{l|}{\textbf{}} & \textbf{1-99\%} & \textbf{2-98\%} & \textbf{5-95\%}  & \textbf{10-90\%}  & \textbf{1-99\%} & \textbf{2-98\%} & \textbf{5-95\%}  & \textbf{10-90\%}  \\
\textbf{HS} & \cellcolor[HTML]{FFFFFD}1.89 & \cellcolor[HTML]{FFFEF7}2.69 & \cellcolor[HTML]{FFFDF0}4.03 & \cellcolor[HTML]{FFFBE1}4.63 & \cellcolor[HTML]{FFFFFA}2.15 & \cellcolor[HTML]{FFFDF2}3.02 & \cellcolor[HTML]{FFFCEB}4.03 & \cellcolor[HTML]{FFFADF}4.41 \\
\textbf{CP} & \cellcolor[HTML]{DBF0E1}1.57 & \cellcolor[HTML]{FFFFFD}2.30 & \cellcolor[HTML]{FFFEF4}3.80 & \cellcolor[HTML]{FFFCE9}4.28 & \cellcolor[HTML]{F0F9F3}1.69 & \cellcolor[HTML]{FFFFFB}2.50 & \cellcolor[HTML]{FFFDF1}3.75 & \cellcolor[HTML]{FFFCE9}3.99 \\
\textbf{IDR}  & \cellcolor[HTML]{FFEF9C}8.57 & \cellcolor[HTML]{FFEF9C}8.63 & \cellcolor[HTML]{FFEF9C}8.49 & \cellcolor[HTML]{FFEF9C}7.70 & \cellcolor[HTML]{FFEF9C}8.41 & \cellcolor[HTML]{FFEF9C}8.43 & \cellcolor[HTML]{FFEF9C}8.15 & \cellcolor[HTML]{FFEF9C}7.05 \\
\textbf{QRA}  & \cellcolor[HTML]{FFF5BD}6.32 & \cellcolor[HTML]{FFF6C2}6.19 & \cellcolor[HTML]{FFF6C3}6.41 & \cellcolor[HTML]{FFF4BA}6.36 & \cellcolor[HTML]{FFF5C0}6.04 & \cellcolor[HTML]{FFF6C4}5.95 & \cellcolor[HTML]{FFF5C1}6.24 & \cellcolor[HTML]{FFF2AF}6.32 \\
\textbf{SQRA} & \cellcolor[HTML]{FFF7CC}5.32 & \cellcolor[HTML]{FFF8D2}5.13 & \cellcolor[HTML]{FFFBE5}4.62 & \cellcolor[HTML]{FFFEF4}3.78 & \cellcolor[HTML]{FFF7CB}5.26 & \cellcolor[HTML]{FFF8D0}5.15 & \cellcolor[HTML]{FFFBE2}4.53 & \cellcolor[HTML]{FFFDED}3.84 \\
\textbf{QRM}  & \cellcolor[HTML]{D0EBD7}1.52 & \cellcolor[HTML]{E4F4E8}1.95 & \cellcolor[HTML]{D9EFDF}2.61 & \cellcolor[HTML]{E5F4E9}2.73 & \cellcolor[HTML]{CBE9D3}1.50 & \cellcolor[HTML]{C6E7CF}1.78 & \cellcolor[HTML]{C2E5CB}2.19 & \cellcolor[HTML]{D6EDDC}2.38 \\
\textbf{SQRM} & \cellcolor[HTML]{81CA95}1.16 & \cellcolor[HTML]{91D1A2}1.40 & \cellcolor[HTML]{A1D7AF}1.71 & \cellcolor[HTML]{A4D9B2}1.38 & \cellcolor[HTML]{73C488}1.06 & \cellcolor[HTML]{73C489}1.19 & \cellcolor[HTML]{7BC88F}1.27 & \cellcolor[HTML]{82CB95}0.88 \\
\textbf{LQRA} & \cellcolor[HTML]{FFFFFD}1.91 & \cellcolor[HTML]{E0F2E4}1.92 & \cellcolor[HTML]{A3D8B1}1.75 & \cellcolor[HTML]{BBE2C6}1.86 & \cellcolor[HTML]{FFFFFE}1.83 & \cellcolor[HTML]{D2ECD9}1.87 & \cellcolor[HTML]{B7E1C2}2.05 & \cellcolor[HTML]{B4DFBF}1.77 \\
\textbf{iQRA} & \cellcolor[HTML]{CAE9D2}1.49 & \cellcolor[HTML]{D1EBD8}1.82 & \cellcolor[HTML]{B1DEBD}1.97 & \cellcolor[HTML]{C0E4C9}1.95 & \cellcolor[HTML]{BEE4C8}1.44 & \cellcolor[HTML]{C6E7CF}1.78 & \cellcolor[HTML]{BDE3C7}2.13 & \cellcolor[HTML]{CBE9D3}2.19 \\
\textbf{iSQRA}  & \cellcolor[HTML]{63BE7B}1.02 & \cellcolor[HTML]{63BE7B}1.09 & \cellcolor[HTML]{63BE7B}0.72 & \cellcolor[HTML]{63BE7B}0.02 & \cellcolor[HTML]{63BE7B}0.98 & \cellcolor[HTML]{63BE7B}1.07 & \cellcolor[HTML]{63BE7B}0.96 & \cellcolor[HTML]{63BE7B}0.32
\end{tabular}
}
\end{table}

\begin{table}[h!]
\centering
\caption{Prediction Interval Pinball Score (PIPS) for the ensemble of 25 XGB point forecasts and four prediction intervals, across two markets.}
\label{tab:PIPS:XGB}
\scalebox{1}{
\begin{tabular}{c|c|c|c|c|c|c|c|c}
\multicolumn{1}{l}{}  & \multicolumn{4}{c|}{\textbf{Germany}}  & \multicolumn{4}{c}{\textbf{France}} \\
\multicolumn{1}{l|}{\textbf{}} & \textbf{1-99\%} & \textbf{2-98\%} & \textbf{5-95\%}  & \textbf{10-90\%}  & \textbf{1-99\%} & \textbf{2-98\%} & \textbf{5-95\%}  & \textbf{10-90\%}  \\
\textbf{HS} & \cellcolor[HTML]{FFFADC}1.02 & \cellcolor[HTML]{FFF2AA}1.60 & \cellcolor[HTML]{FFF0A0}2.84 & \cellcolor[HTML]{FFF0A2}4.22 & \cellcolor[HTML]{FFFCE7}1.03 & \cellcolor[HTML]{FFF8D3}1.64 & \cellcolor[HTML]{FFF1A9}2.97 & \cellcolor[HTML]{FFF2AA}4.46 \\
\textbf{CP} & \cellcolor[HTML]{FFFADB}1.02 & \cellcolor[HTML]{FFF1A6}1.61 & \cellcolor[HTML]{FFEF9C}2.85 & \cellcolor[HTML]{FFEF9C}4.24 & \cellcolor[HTML]{FFFBE6}1.03 & \cellcolor[HTML]{FFF8D0}1.65 & \cellcolor[HTML]{FFEF9C}3.00 & \cellcolor[HTML]{FFEF9C}4.50 \\
\textbf{IDR}  & \cellcolor[HTML]{FFEF9C}1.19 & \cellcolor[HTML]{FFEF9C}1.63 & \cellcolor[HTML]{FFF8D0}2.69 & \cellcolor[HTML]{FFFCE9}3.98 & \cellcolor[HTML]{FFEF9C}1.30 & \cellcolor[HTML]{FFEF9C}1.80 & \cellcolor[HTML]{FFF4B7}2.93 & \cellcolor[HTML]{FFFADB}4.31 \\
\textbf{QRA}  & \cellcolor[HTML]{FFFBE1}1.00 & \cellcolor[HTML]{FFFBE5}1.47 & \cellcolor[HTML]{FFFDEE}2.59 & \cellcolor[HTML]{FFFCEB}3.97 & \cellcolor[HTML]{FFF7CB}1.13 & \cellcolor[HTML]{FFF8D3}1.64 & \cellcolor[HTML]{FFF7CD}2.87 & \cellcolor[HTML]{FFF7CB}4.35 \\
\textbf{SQRA} & \cellcolor[HTML]{FFFADD}1.02 & \cellcolor[HTML]{FFFBE1}1.48 & \cellcolor[HTML]{FFFDF2}2.58 & \cellcolor[HTML]{FFFDF0}3.95 & \cellcolor[HTML]{FFF7C9}1.14 & \cellcolor[HTML]{FFF8D2}1.65 & \cellcolor[HTML]{FFF8D3}2.85 & \cellcolor[HTML]{FFF8D3}4.33 \\
\textbf{QRM}  & \cellcolor[HTML]{B8E1C3}0.84 & \cellcolor[HTML]{D3ECDA}1.35 & \cellcolor[HTML]{E6F4EA}2.49 & \cellcolor[HTML]{E3F3E7}3.85 & \cellcolor[HTML]{91D1A2}0.85 & \cellcolor[HTML]{98D4A7}1.39 & \cellcolor[HTML]{B2DFBE}2.61 & \cellcolor[HTML]{BCE3C6}4.07 \\
\textbf{SQRM} & \cellcolor[HTML]{B5E0C1}0.84 & \cellcolor[HTML]{D0EBD8}1.34 & \cellcolor[HTML]{E4F3E8}2.49 & \cellcolor[HTML]{DFF2E4}3.84 & \cellcolor[HTML]{86CC98}0.84 & \cellcolor[HTML]{91D1A2}1.38 & \cellcolor[HTML]{AFDDBB}2.61 & \cellcolor[HTML]{B9E2C4}4.07 \\
\textbf{LQRA} & \cellcolor[HTML]{63BE7B}0.74 & \cellcolor[HTML]{63BE7B}1.19 & \cellcolor[HTML]{63BE7B}2.26 & \cellcolor[HTML]{63BE7B}3.61 & \cellcolor[HTML]{94D2A4}0.85 & \cellcolor[HTML]{68C07F}1.33 & \cellcolor[HTML]{63BE7B}2.48 & \cellcolor[HTML]{63BE7B}3.91 \\
\textbf{iQRA} & \cellcolor[HTML]{7BC88F}0.77 & \cellcolor[HTML]{87CD9A}1.24 & \cellcolor[HTML]{83CB96}2.32 & \cellcolor[HTML]{74C589}3.64 & \cellcolor[HTML]{72C487}0.82 & \cellcolor[HTML]{6AC181}1.33 & \cellcolor[HTML]{6EC284}2.50 & \cellcolor[HTML]{6EC285}3.93 \\
\textbf{iSQRA}  & \cellcolor[HTML]{77C68C}0.77 & \cellcolor[HTML]{82CB95}1.23 & \cellcolor[HTML]{7FC992}2.31 & \cellcolor[HTML]{6FC385}3.63 & \cellcolor[HTML]{63BE7B}0.81 & \cellcolor[HTML]{63BE7B}1.32 & \cellcolor[HTML]{69C080}2.50 & \cellcolor[HTML]{6AC181}3.93
\end{tabular}
}
\end{table}

\begin{figure}[h!]
\caption{{Gain per trade (EUR) and loss avoidance (\%) from strategies based on postprocessed ensemble of 25 XGB forecasts, on the German market, aggregated for the entire sample period. Baseline profit per trade equates to 40.1 EUR and is marked by a dashed line. The bars represent strategies with different risk thresholds - $1/3$ for dark blue, $2/3$ for blue, and $1$ for light blue color.}}
\label{fig:Trading:XGB}
\includegraphics[width=\textwidth]{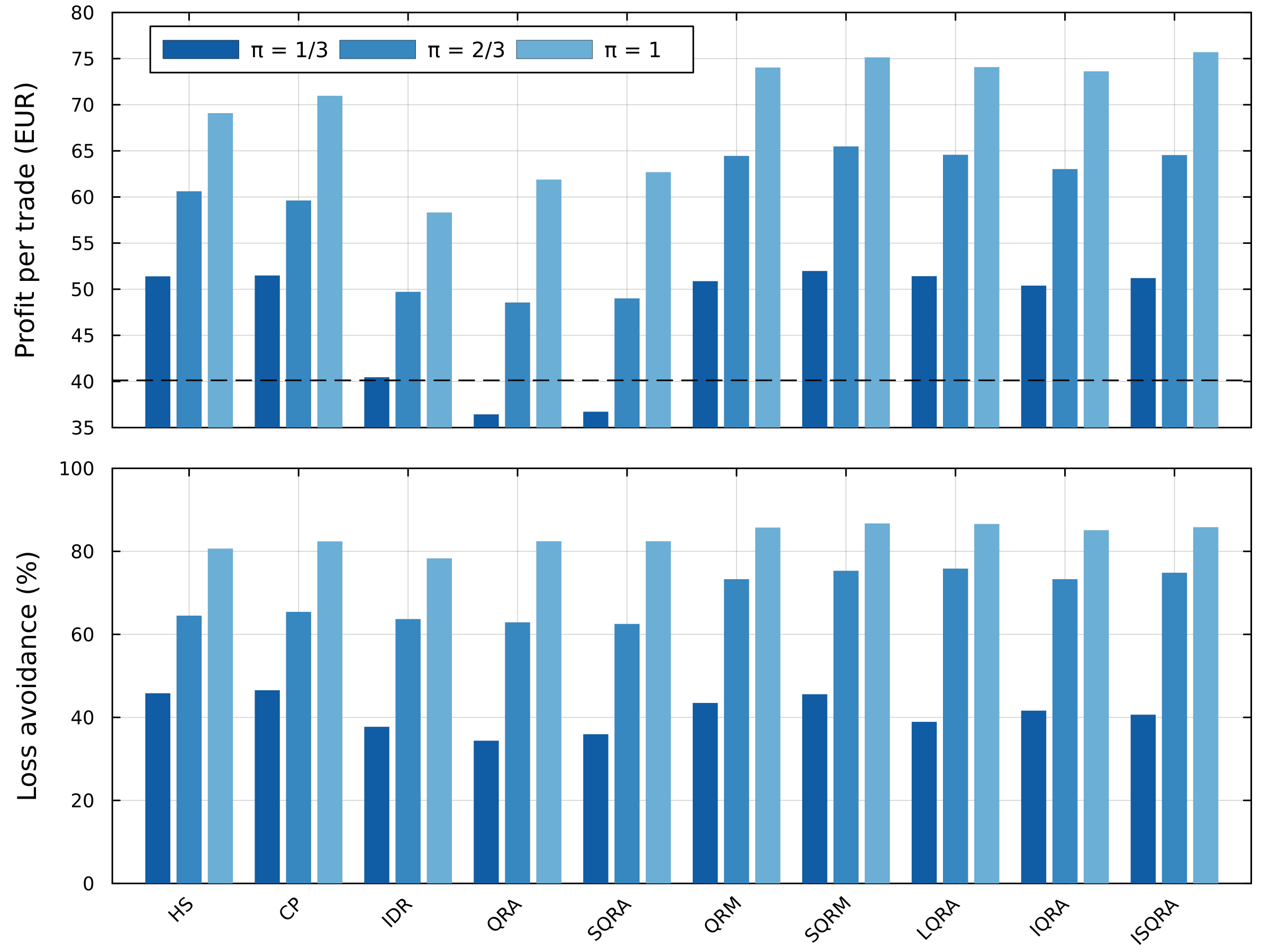}
\end{figure}

\begin{table}[]
\caption{{Number of trades and profit per trade for the economic evaluation of the postprocessed ensemble of 25 XGB models in the German day-ahead market. Profit per trade corresponds to the values plotted in Fig.~\ref{fig:Trading:XGB}}.}
\centering
\begin{tabular}{c|c|c|c|c|c|c}
\multicolumn{1}{c}{} & \multicolumn{3}{c|}{\textbf{Number of trades}} & \multicolumn{3}{c}{\textbf{Profit per trade}} \\
$\pi$ & $1/3$ & $2/3$ & $1$ & $1/3$ & $2/3$ & $1$ \\
\hline
\textbf{HS}                   & 522           & 409           & 317           & 51.4          & 60.6          & 69.1          \\
\textbf{CP}                   & 519           & 409           & 306           & 51.5          & 59.6          & 71.0          \\
\textbf{IDR}                  & 587           & 472           & 340           & 40.5          & 49.7          & 58.3          \\
\textbf{QRA}                  & 597           & 452           & 316           & 36.4          & 48.6          & 61.9          \\
\textbf{SQRA}                 & 591           & 449           & 316           & 36.7          & 49.0          & 62.7          \\
\textbf{QRM}                  & 513           & 376           & 269           & 50.9          & 64.4          & 74.0          \\
\textbf{SQRM}                 & 507           & 371           & 262           & 52.0          & 65.5          & 75.1          \\
\textbf{LQRA}                 & 489           & 355           & 245           & 51.4          & 64.6          & 74.1          \\
\textbf{iQRA}                 & 512           & 369           & 266           & 50.4          & 63.0          & 73.6          \\
\textbf{iSQRA}                & 506           & 361           & 258           & 51.2          & 64.5          & 75.7         
\end{tabular}
\end{table}

\end{document}